\pdfoutput=1
\documentclass[10pt,twocolumn,letterpaper]{article}

\usepackage[letterpaper,textwidth=7.0in,textheight=9.0in,columnsep=0.375in,
            centering,top=0.75in]{geometry}

\usepackage[T1]{fontenc}
\usepackage[utf8]{inputenc}
\usepackage{newtxtext,newtxmath}   

\usepackage[hyphens]{url}
\usepackage{graphicx}
\usepackage{caption}
\usepackage{placeins}              
\usepackage{natbib}
\let\cite\citep
\setcitestyle{aysep={}}

\usepackage{booktabs}
\usepackage{tabularx}
\usepackage{colortbl}   
\usepackage{array}
\usepackage{multirow}
\usepackage{tikz}
\usetikzlibrary{arrows.meta,positioning,matrix}
\definecolor{kcbgood}{HTML}{DDEEDF}
\definecolor{kcbbad}{HTML}{F4DCDC}
\definecolor{kcbgoodlight}{HTML}{EFF7F0}
\definecolor{kcbbadlight}{HTML}{FBECEC}
\newcommand{\kcbgoodholm}[1]{\cellcolor{kcbgood}\textbf{$#1$}}
\newcommand{\kcbbadholm}[1]{\cellcolor{kcbbad}\textbf{$#1$}}
\newcommand{\kcbgoodnom}[1]{\cellcolor{kcbgoodlight}$#1^{\dagger}$}
\newcommand{\kcbbadnom}[1]{\cellcolor{kcbbadlight}$#1^{\dagger}$}
\newcommand{\kcbnull}[1]{$#1$}
\usepackage{xspace}
\newcommand{\dataset}{MemToC\xspace}

\usepackage{microtype}
\newcommand{\FLAG}[1]{}
\providecommand{\tightlist}{}

\begin{document}

\twocolumn[%
\begin{center}
{\LARGE\bfseries MemToC: Benchmarking Memory--Tool\\[0.25em]
Conflict Resolution in Large Language Models\par}

\vspace{1.4em}

{\large
Arseniy Varlamov\textsuperscript{1} \quad
Rishat Zinnatullin\textsuperscript{2} \quad
Elisei Rykov\textsuperscript{3} \quad
Alexander Panchenko\textsuperscript{3,4} \quad
Ilseyar Alimova\textsuperscript{3}\par}

\vspace{0.7em}

{\small
\textsuperscript{1}Central University, Moscow \quad
\textsuperscript{2}Ural Federal University, Yekaterinburg\\
\textsuperscript{3}Skolkovo Institute of Science and Technology, Moscow \quad
\textsuperscript{4}AIRI, Moscow\par}
\end{center}

\vspace{1.6em}

\begin{center}
\begin{minipage}{0.86\textwidth}
{\centering\large\bfseries Abstract\par}
\vspace{0.5em}
\noindent Tool-augmented LLMs must arbitrate between two fallible sources when a tool return conflicts with their parametric memory, yet existing evaluations often measure source preference without establishing source correctness. We introduce \dataset, a controlled benchmark for post-tool-return arbitration with executable tools. \dataset comprises 6,504 evaluation episodes constructed from 542 quality-controlled factual questions, independently elicited model-specific closed-book answers, and controlled tool returns of known correctness. These components instantiate four source-correctness cases, while tool-error and no-tool conditions serve as separate controls. The resulting metrics distinguish appropriate source use from indiscriminate source preference. Across five open-weight 7--9B models, tool returns strongly dominate elicited closed-book answers. The four instruction-tuned models retain a verified-correct answer against an incorrect tool in only 6.5--17.1\% of eligible cases, follow a correct tool in 86.0--93.1\%, and repeat the tool return in 78.4--86.0\% of cases where both sources are wrong. No cross-model ordering remains stable across three instruction-wording variants with the question and episode content held fixed. Motivated by these failures, we compare prompting with SFT and DPO using chain-level cross-fitting over ToolHop, ensuring that questions sharing an underlying fact never straddle training and evaluation. We apply an asymmetric success criterion: correct-answer retention must improve without a detected reduction in correct-tool following. SFT and DPO meet this criterion on the same two of four instruction-tuned backbones, suggesting that outcomes depend more strongly on the backbone than on the optimization objective within our model set. Improvements rarely come cleanly: 19 of 20 tested method--model combinations reduce abstention after tool errors or on unanswerable inputs. Transfer beyond \dataset is positive but partial and depends on the model and presentation frame. Thus, correctness-conditioned arbitration can be improved through fine-tuning, but gains must be evaluated jointly with correct tool use, abstention, and robustness to formulation.

\end{minipage}
\end{center}

\vspace{2.0em}
]

\providecommand{\kcbstep}[2]{%
  \par\smallskip\noindent\textbf{#1. #2}%
  \par\nobreak\noindent\ignorespaces}

\providecommand{\kcbtarget}{0}%

\providecommand{\kcbsec}[2]{\ifcsname r@#1\endcsname\ref{#1}\else#2\fi}

\ifnum\kcbtarget=2\relax\else%
\begin{figure}[t]
    \centering
    \includegraphics[width=\columnwidth]{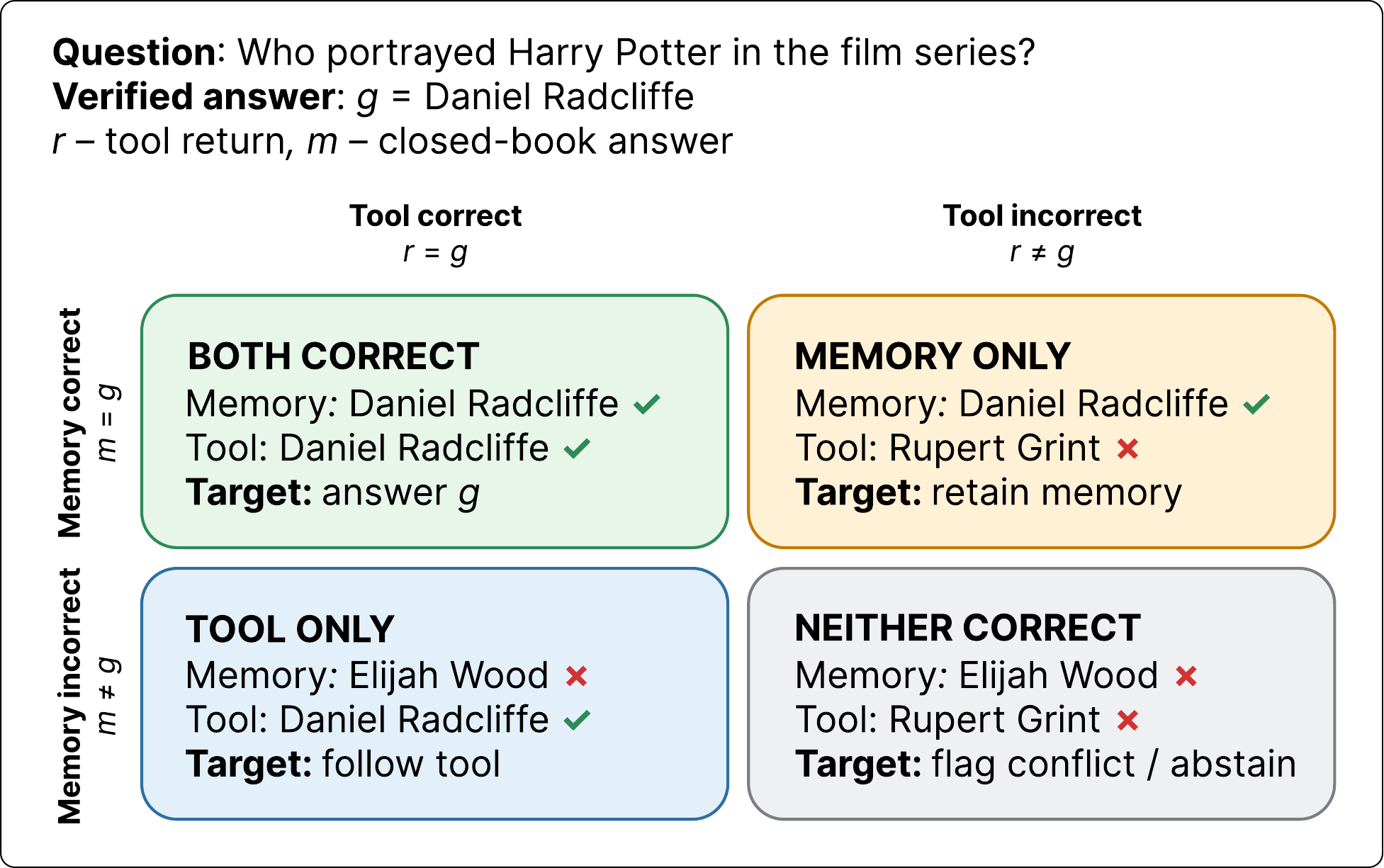}
    \caption{Schematic illustration of the four source-correctness cases. The displayed combination of answer values is hypothetical and does not represent a single observed \dataset{} episode; a fixed preference for either source cannot succeed across all four cases.}
    \label{fig:motivation}
\end{figure}
\fi%

\ifnum\kcbtarget=3\relax\else%
\section{Introduction}

LLM agents increasingly rely on tools whose outputs are presented as authoritative evidence, yet tools can be stale, misconfigured, or resolve a query to the wrong entity. When a tool return disagrees with a model's parametric memory, the system must arbitrate between two fallible sources, and evaluating that requires knowing which source is correct: a metric recording only which source is followed cannot separate useful from harmful deference (Figure~\ref{fig:motivation}).

Existing work lacks explicit control over source correctness. TMC~\cite{cheng2026investigatingtoolmemoryconflictstoolaugmented} identifies conflicts as disagreements between memory-only and tool-conditioned responses. Because these conflicts emerge naturally rather than being constructed, neither source correctness nor the appropriate arbitration decision is known in advance. Other work constructs conflicts through entity substitution~\cite{longpre-etal-2021-entity} or counterfactual augmentation~\cite{neeman-etal-2023-disentqa}, and recent benchmarks stratify textual evidence by source correctness~\cite{sun2026pave,xie2024adaptive}. However, these settings do not study executable tool returns and omit tool-error and no-tool controls. Agentic evaluations likewise show models overwhelmingly adopting tool outputs despite access to reliability diagnostics~\cite{whenthetooldecides2026}. We instead study the controlled post-tool-call setting, where tool selection and invocation are held fixed while only the observed tool return is intervened upon.

\paragraph{Our approach}
We construct \dataset from the ToolHop~\cite{ye-etal-2025-toolhop} dataset, which provides decomposed multi-hop questions, locally executable tool calls, and verifiable intermediate answers. From these data, we derive a quality-controlled benchmark set of self-contained factual questions with short, verifiable answers and plausible same-type distractors. For each model and question, we first elicit a closed-book answer and then present the same question with a correct tool return, an incorrect return, an error payload, or no return. Tool-return correctness is controlled by injecting either a correct or an incorrect return constructed from externally verified answers and human-validated distractors, whereas LLM memory correctness is measured independently for each model by comparing its closed-book answer against the verified answer. Because we supply the return rather than observe it, we know whether the tool is right before the model answers, instead of having to judge it afterwards. Together with the separately measured closed-book answer, this fixes the correct response for every episode, which is what makes arbitration --- rather than source preference --- measurable. Tool-error and no-tool instances serve as separate controls. This design evaluates resistance to incorrect returns separately from appropriate use of correct returns, rather than rewarding a fixed preference for either source.
We make three contributions.
\begin{enumerate}
    \item \textbf{Benchmark.} We introduce \dataset, a controlled benchmark for tool--memory arbitration comprising 542 quality-controlled factual questions with executable calls, verified answers, model-specific closed-book answers, and controlled correct and incorrect returns.
    \item \textbf{Evaluation.} We evaluate five open-weight 7--9B models across four source-correctness cases, complemented by tool-error and no-tool controls, extraction-first answer matching, and three instruction-wording variants with fixed question text.
    \item \textbf{Mitigation methods.} We compare prompting with SFT and DPO using cross-fitting over ToolHop chains, so that every reported fine-tuned prediction is obtained on questions held out from training.
\end{enumerate}
\fi%

\ifnum\kcbtarget=2\relax\else%

\ifnum\kcbtarget=3\relax\subsection{Related work}\else\section{Related Work}\fi\label{sec:related}

\paragraph{Tool--memory conflicts}
TMC~\cite{cheng2026investigatingtoolmemoryconflictstoolaugmented} introduced tool--memory conflicts by identifying disagreements between memory-only and tool-conditioned responses, and showed that existing mitigation strategies only partially resolve them. Subsequent work finds that models may defer strongly to tool outputs and that failures can arise from the tool-use protocol itself~\cite{whenthetooldecides2026,zhang2026tooluse}. Concurrent studies examine tool reliability and failure modes directly~\cite{tian2026toolbenchx,zhang2024toolbehonest,yan2026trustnotool,zhu2026whentoolsfail,soni2026toolfailbench}. However, these studies do not independently establish the correctness of both the model's closed-book answer and the tool return, and therefore cannot evaluate arbitration across source-correctness cases.

\paragraph{Controlling source preference} Direct Preference Optimization (DPO)~\cite{dpo} learns from preferred and rejected responses without training an explicit reward model. Context-DPO~\cite{bi-etal-2025-context} applies this objective to following contextual evidence, while KnowPO~\cite{zhang2025knowpo} learns adaptive source selection over constructed conflicts. Both methods primarily optimize reliance on contextual information rather than condition source selection on the correctness of competing sources. Correctness-aware methods in RAG use gating, regime-specific models, or conflict-aware decoding~\cite{zhu2026saber,wang2026rapsda,jiang2026conflictaware}. Other approaches steer models between parametric memory and contextual evidence~\cite{zhao2024spare,xin-etal-2025-sparse,li2026shift}, consolidate evidence at inference time~\cite{wang2025astute}, or control whether a tool should be invoked~\cite{wang2026asa,chen2026headingsteering,zeng2026tooloveruse}. None of these methods evaluates correctness-conditioned arbitration after a tool return. 

\ifnum\kcbtarget>1\relax
\begin{figure}[p]
    \centering
    \includegraphics[trim=0 0 2060 0,clip,width=\textwidth]{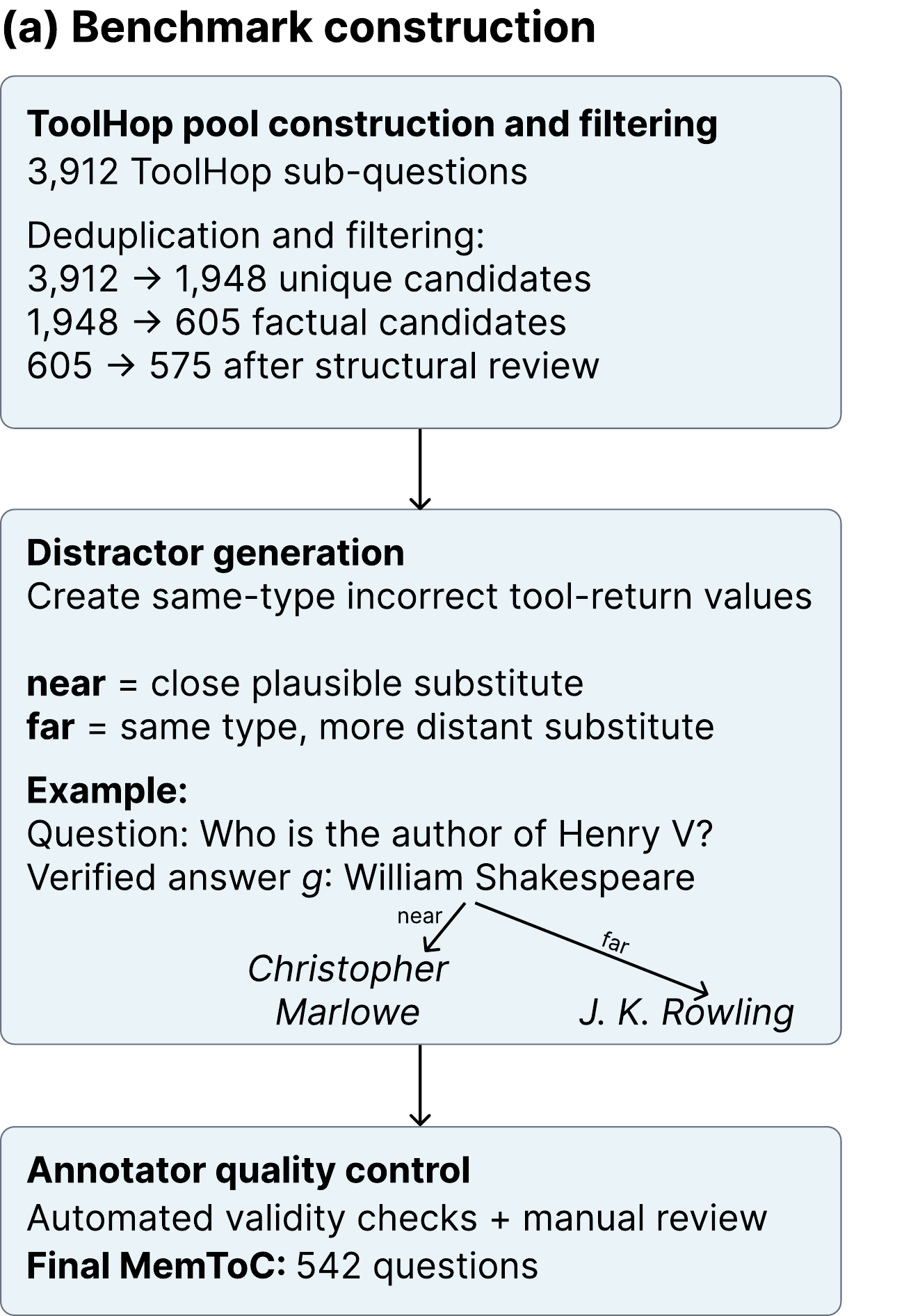}
    \caption{\textbf{Overview of \dataset, part (a): benchmark construction.}
ToolHop sub-questions are filtered, augmented with type-consistent near and
far incorrect returns, and quality-controlled to yield 542 factual items.}
    \label{fig:kcb-overview}
\end{figure}
\begin{figure}[t]
    \centering
    \includegraphics[trim=635 0 0 0,clip,width=\textwidth]{Figures/benchmark.pdf}
    \caption{\textbf{Overview of \dataset, part (b): model-specific evaluation.}
For each model and prompt formulation, a closed-book pass obtains $m$,
followed by a tool-conditioned pass that produces $y$ under controlled return
$r$. Comparing $m$ and $r$ with verified answer $g$ stratifies responses into
four source-correctness cases, separating answer retention from correct- and
incorrect-tool following; control conditions are evaluated separately.}
    \label{fig:kcb-overview-b}
\end{figure}
\else
\begin{figure*}[htb!]
    \centering
    \includegraphics[width=\textwidth]%
        {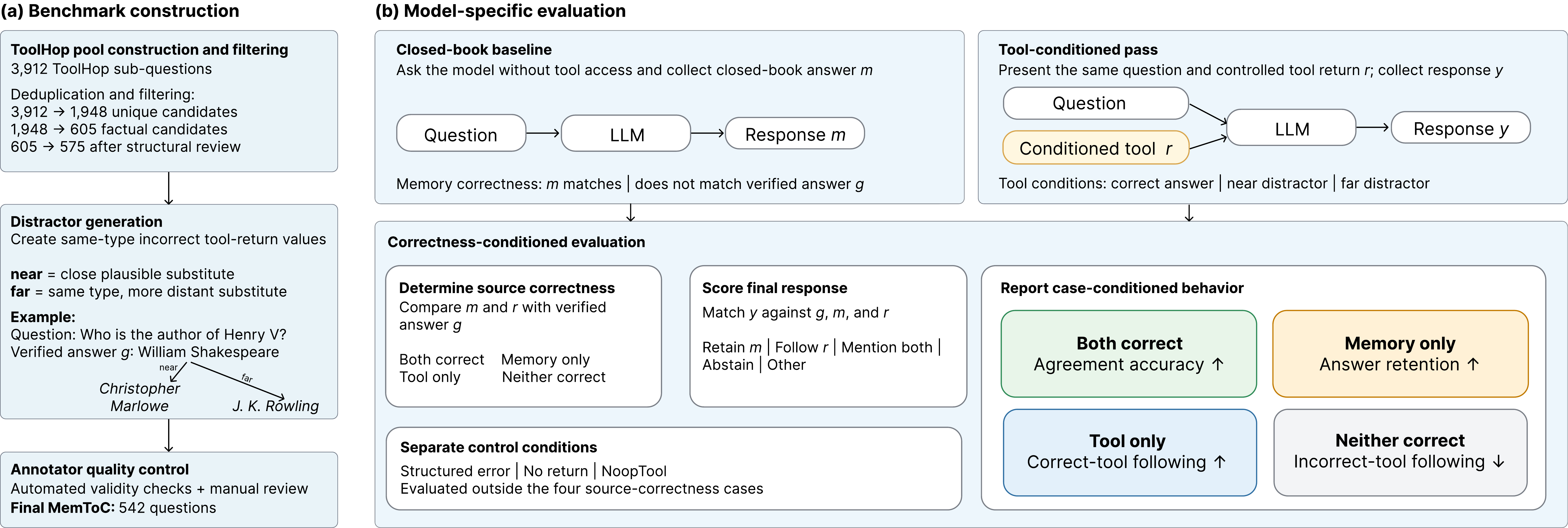}
    \caption{\textbf{Overview of \dataset.} (a) ToolHop sub-questions are filtered, augmented with type-consistent near and far incorrect returns, and quality-controlled to yield 542 factual items. (b) For each model and prompt formulation, a closed-book pass obtains $m$, followed by a tool-conditioned pass that produces $y$ under controlled return $r$. Comparing $m$ and $r$ with verified answer $g$ stratifies responses into four source-correctness cases, separating answer retention from correct- and incorrect-tool following; control conditions are evaluated separately.}
    \label{fig:kcb-overview}
\end{figure*}
\fi
\fi%

\ifnum\kcbtarget=3\relax\else%
\section{The \dataset Benchmark}
\label{sec:kcb-benchmark}

\dataset separates source correctness from source preference (Figure~\ref{fig:kcb-overview}): questions, verified answers, executable calls, and controlled incorrect returns are constructed and quality-controlled before any model is evaluated. The closed-book answer is then elicited separately for each model and prompt formulation, making case membership model- and formulation-specific while the benchmark artifacts remain fixed. Each benchmark question includes a verified answer, an executable tool call, and at least one curated incorrect return; combining 542 quality-controlled questions with four tool conditions and three instruction-wording variants yields 6,504 fixed evaluation episodes.

\subsection{Source data}
\label{sec:kcb-source}

\dataset is built from ToolHop~\cite{ye-etal-2025-toolhop}, whose 995 chains decompose into 3,912 sub-questions, each with an executable tool call and a verified intermediate answer. We use the sub-questions and discard each chain's terminal answer, which in 602 of the 995 chains is computed over earlier steps rather than being an independently queryable fact. The retained questions have short, verifiable answers spanning named entities such as people, places, organizations, and creative works, temporal values such as dates, years, and time zones, and other short textual values.

\subsection{Construction pipeline}
\label{sec:kcb-construction}

\fi%

\ifnum\kcbtarget=1
Eight steps turn ToolHop sub-questions into \dataset records. Steps~2, 4 and 7 are \emph{manual}, Steps~1, 5 and 6 \emph{automatic}, Step~3 both, and \textbf{every manual step uses two annotators working independently}. Rule sets, constants and per-branch audits are in Appendix~\kcbsec{app:d}{D}.

\kcbstep{Step 1 (auto)}{Flatten chains and deduplicate.}
The unit is the sub-question, not the chain step: one formulation recurs in 17 different chains, so step-level sampling yields far less variety than its size suggests. \textbf{3,912 $\rightarrow$ 1,948 unique.}

\kcbstep{Step 2 (manual+auto)}{Keep only factual questions.}
An answer that is computed rather than recalled cannot conflict with parametric knowledge, so a rule-based filter rejects answers produced by an operation over an earlier one. Because such a filter over-rejects, two annotators reviewed the 54 unclear rejections and \textbf{restored 19} --- a city's time zone is a fact about a place, not an arithmetic result. \textbf{1,948 $\rightarrow$ 605.}

\kcbstep{Step 3 (auto+manual)}{Control redundancy and leakage.}
Two questions sharing an underlying fact must not be split across folds. Automatic passes identify near-identical formulations and \emph{knowledge clusters} --- linked questions about one family or narrative --- and two annotators adjudicated the 58 ambiguous cases. Nothing is removed; the output is the grouping used for fold assignment. \textbf{605 $\rightarrow$ 605}, grouped.

\kcbstep{Step 4 (manual)}{Structural review of wording.}
Each retained question must stand alone and not give away its answer. \textbf{We did not rewrite or regenerate any question: we preserved the original ToolHop wording, and self-containedness was a selection criterion rather than an editing operation.} Two annotators independently reviewed all 605 candidates and removed 30; verified answers were checked against Wikidata, by a web-browsing language model, and manually against primary sources. \textbf{605 $\rightarrow$ 575.}

\kcbstep{Step 5 (auto+manual)}{Answer-aware distractor construction.}
A credible incorrect value must match the question's domain and period as well as its answer type, which a draw from ToolHop's own pool --- another question's verified answer of the same nominal type --- does not deliver. Each substitution is built from the verified entity itself, through a \emph{Wikidata} branch (246 questions, matching on occupation, gender, birth year and citizenship), a \emph{temporal} branch (173, offsetting dates and years deterministically in the verified answer's own format), or a \emph{human-authored} branch (149, where Wikidata cannot resolve the entity). All preserve the answer type: a \emph{near} distractor is indistinguishable from the verified answer in type and plausibility, a \emph{far} one obviously different in period or country, with the branch-specific realization frozen in the released mapping. With seven repairs, \textbf{575 $\rightarrow$ 575 records}, 426 automatic and 149 authored.

\kcbstep{Step 6 (auto)}{Programmatic validity checks.}
Four assertions in the build script, no human judgement, catch malformed records: every distractor non-empty, none equal to the verified answer, near and far distinct, none coinciding with an alternative true value Wikidata holds for that property. All pass. \textbf{575 $\rightarrow$ 575.}

\kcbstep{Step 7 (manual)}{Blinded semantic review.}
Code cannot check whether a question makes sense alone or whether its incorrect value is believable. Two annotators reviewed all 575 records under five such checks, blind to each other: 503 \textsc{pass}, 39 \textsc{repair}, 33 \textsc{exclude}, a \textsc{repair} changing only the substituted value. \textbf{575 $\rightarrow$ 542.}

\kcbstep{Step 8 (auto)}{Instantiation into tool conditions.}
This step turns each record into the episodes a model sees, so that only the observed return varies between conditions. Each of the \textbf{542} questions is instantiated with a correct return carrying $g$, an incorrect return carrying $d_{\mathrm{near}}$, one carrying $d_{\mathrm{far}}$ where available, a structured error payload, and no return, each under three instruction-wording variants with fixed question text; \emph{NoopTool} is a separate interface control (Appendix~\kcbsec{app:a}{A}).

\paragraph{Why part of the distractor layer is authored} The share of authored records is a measured outcome, not a stylistic choice: blinded sampled audits leave the automatic branch with a residual rejection rate of about 12\% against a target of 3\%, concentrated on pre-modern and legendary figures whose Wikidata records carry no citizenship, leaving the candidate query culturally unconstrained. We read this as the ceiling of the automatic method and route the residue to human authors.

\paragraph{Composition} The 605 factual questions come from 310 ToolHop chains and the 575 retained after Step~4 from 309 of them; a question occurring in several chains is assigned to the first, so the remaining chains are not free of factual questions --- their texts collapsed during deduplication. The corpus is long-tailed and biographical --- 47 topics, the ten largest carrying 439 of the 575 questions (76.3\%), led by history, film and genealogy --- which is why people dominate the answer types, 304 of 575, and why the Wikidata branch covers most of the pool. A far distractor exists for 463 of the 575 questions (443 after the Step~7 exclusions), missing for most authored records and for all \texttt{other\_string} and \texttt{work} answers, so near--far comparisons run on a subset skewed towards automatically constructed items (Appendix~\kcbsec{app:d}{D}).
\else
\ifnum\kcbtarget=2\relax\else%
\ifnum\kcbtarget=3\relax\subsection{Construction pipeline: candidate extraction}\fi
The left side of Figure~\ref{fig:kcb-overview} summarizes the construction pipeline in three stages. All manual reviews and authoring decisions were performed independently by two annotators.

\paragraph{Candidate extraction and filtering} We flatten ToolHop's 3,912 sub-questions and deduplicate identical formulations, yielding 1,948 unique candidates. We retain 605 factual questions whose answers can be retrieved independently rather than computed from earlier chain steps. Related and near-duplicate questions are grouped for fold assignment, after which structural and answer verification removes 30 questions that are not self-contained, reveal their answer, or have an invalid verified answer. This produces a 575-question construction pool.

\fi%

\ifnum\kcbtarget=3\relax\else%

\paragraph{Distractor construction} Each question receives a plausible \emph{near} distractor and, where possible, a \emph{far} distractor. Near distractors match the verified answer in type, domain, and plausibility, whereas far distractors preserve the answer type but differ more strongly in period or country. Distractors are generated automatically for 426 records using Wikidata constraints or deterministic temporal transformations. For the remaining 149 records, where structured metadata is insufficient to identify a plausible substitute, two annotators author the distractors. The size of this share is a measured outcome rather than a stylistic choice: blinded audits of sampled records leave the automatic branch with a residual rejection rate of about 12\% against a target of 3\%, concentrated on pre-modern and legendary figures whose Wikidata records carry no citizenship. Both automatically generated and human-authored distractors subsequently undergo the same blinded quality-control procedure.

\paragraph{Quality control} Programmatic checks require every distractor to be non-empty, distinct from the verified answer and from the paired distractor, and absent from Wikidata's alternative valid values for the queried property. Two annotators then review all 575 records for self-containedness, answer leakage, verified-answer validity, distractor type, and plausibility. The review yields 503 passes, 39 distractor repairs, and 33 exclusions, producing the final 542-question benchmark. After quality control, far distractors are available for 443 of the 542 benchmark questions. Their coverage is lower among records with human-authored distractors and absent for creative works and other short textual values. Consequently, near--far comparisons are restricted to a non-random subset weighted toward automatically constructed distractors.
\fi%
\fi

\ifnum\kcbtarget=3\relax\else%
\subsection{Evaluation procedure}
\label{sec:kcb-evaluation}

\ifnum\kcbtarget>1\relax Figure~\ref{fig:kcb-overview-b} summarizes the evaluation procedure in three stages.\else Figure~\ref{fig:kcb-overview}(b) summarizes the evaluation procedure in three stages.\fi

\paragraph{Closed-book baseline} For each model and prompt formulation, we first elicit an answer $m$ without tool access. The prompt requires the model to provide its best guess rather than abstain, ensuring that $m$ can be evaluated against the verified answer $g$. The correctness of $m$ determines the row of the source-correctness matrix.

\paragraph{Tool-conditioned run} We then present the same question together with a controlled tool return $r$ and collect the model's final response $y$. The question text, verified answer, executable call, and return are fixed benchmark artifacts, while $m$ and $y$ are model-specific. Each item is evaluated under a reference prompt and two wording variants of the closed-book and with-tool instructions. Return conditions are a correct return, a near incorrect return, a structured error payload, and no return. A far incorrect return is evaluated separately, on the 443 questions that have one. \texttt{NoopTool} serves as a separate interface control that preserves the tool-call format without providing a competing factual value.

\paragraph{Correctness-case assignment} The correctness of $m$ and $r$ relative to $g$ assigns each model--formulation pair to one of four cases. When both sources are correct, the target is to return $g$. When only the closed-book answer is correct, the target is to retain $m$. When only the tool return is correct, the target is to follow $r$. When neither source is correct, the target is to reject both answers, explicitly surface the conflict, and abstain. Structured tool errors, no-return instances, and \texttt{NoopTool} are evaluated as controls outside this four-case matrix. Case assignment is completed before the final response $y$ is scored.

\fi%

\ifnum\kcbtarget=2\relax\else%
\subsection{Validation of response targets}
\label{sec:kcb-validation}

To validate the response targets independently of the benchmark labels, two annotators labeled the preferred assistant action in 100 conflict scenarios without seeing the assigned source-correctness case, reaching four-class inter-annotator agreement (Cohen’s $\kappa=0.785$). The annotators generally preferred responses that explicitly surfaced a disagreement rather than silently selecting one source. When neither source was correct, they preferred an explicit statement of uncertainty to an unexplained refusal. These annotations support treating conflict acknowledgment as a separate outcome and flagged abstention as the target when neither source is correct.
\fi%

\ifnum\kcbtarget=3\relax\else%
\section{Experimental Setup}
\label{sec:experimental-setup}

We evaluate arbitration behavior on five open-weight LLMs using the proposed benchmark and a preregistered evaluation protocol. Beyond measuring baseline performance, we investigate two classes of mitigation methods: prompting and parameter-efficient fine-tuning, and assess the robustness of the resulting behavior to changes in prompt formulation, presentation format, and evaluation setting. Unless stated otherwise, all experiments follow the same inference and scoring protocol described below.

\subsection{Models and inference}

We evaluate five open-weight models: Llama-3.1-8B\footnote{\url{https://hf.co/meta-llama/Llama-3.1-8B}}, Llama-3.1-8B-Instruct\footnote{\url{https://hf.co/meta-llama/Llama-3.1-8B-Instruct}}, Qwen2.5-7B-Instruct\footnote{\url{https://hf.co/Qwen/Qwen2.5-7B-Instruct}}, gemma-2-9b-it\footnote{\url{https://hf.co/google/gemma-2-9b-it}}, and Mistral-7B-Instruct-v0.3\footnote{\url{https://hf.co/mistralai/Mistral-7B-Instruct-v0.3}}. These models cover four families and include a direct comparison between the base and instruction-tuned Llama checkpoints. All models are served with vLLM on a single NVIDIA A100 80GB GPU and decoded greedily at temperature 0.

\fi%

\ifnum\kcbtarget=2\relax\else%
\subsection{Evaluation implementation and metrics}
\label{sec:kcb-scoring}

\paragraph{Closed-book prompt sensitivity} The primary evaluation uses the required-best-guess prompt defined in the benchmark protocol. As a sensitivity analysis, we also evaluate a prompt that allows the model to respond that it does not know. Allowing abstention leaves some source-correctness cases nearly empty for refusal-prone models, so the required-best-guess condition is used for the primary analysis. Retention under this condition should therefore be interpreted as an upper bound on the model's ability to retain a correct closed-book answer under ordinary interaction.

\paragraph{Response scoring} We request responses in the format \texttt{FINAL: <answer>} and extract the marked span, falling back to the final line when the marker is absent. Normalization- and alias-tolerant matching compares the extracted answer with the verified answer, the closed-book answer, and the observed tool return. Responses are classified as following the tool, retaining the closed-book answer, mentioning both, abstaining, or selecting another answer. The same extraction and matching procedure is applied to closed-book and tool-conditioned responses. For predefined ambiguous abstentions that cannot be resolved by the rule-based parser, a fixed LLM-based extractor returns a normalized span to the same matcher. Against blinded human labels on 167 responses, the deterministic scorer reaches $\kappa=0.806$, with recall 1.00 for tool following, 0.939 for retention, and 0.706 for abstention. The LLM-based normalization is restricted to these ambiguous abstentions and changes retention by at most 3.1 percentage points and incorrect-tool following by at most 0.6 percentage points.

\paragraph{Metrics} Correct-answer retention measures how often the model returns the verified answer when its closed-book answer is correct and the tool return is incorrect. Correct-tool following measures how often it follows a verified-correct return when its closed-book answer is incorrect. Conflict-resolved accuracy measures verified-answer accuracy across these two cases. Incorrect-tool following measures adoption of the tool return when neither source is correct, and tool-error abstention measures abstention after a structured tool error. Higher values are desirable for all metrics except incorrect-tool following. Agreement-case accuracy and conflict acknowledgment are reported as secondary metrics.

\fi%

\ifnum\kcbtarget=3\relax\else%
\subsection{Mitigation methods}
\label{sec:mitigation}

\paragraph{Success criterion} We consider a mitigation method successful if correct-answer retention improves significantly and no statistically significant reduction in correct-tool following is detected. This criterion rules out methods that resist incorrect returns only by inducing general distrust of tool outputs. All comparisons use common question support.
\fi%

\ifnum\kcbtarget=2\relax\else%
\ifnum\kcbtarget=3\relax\subsection{Mitigation methods (continued)}\fi
\paragraph{Prompting} We evaluate three prompting strategies on the four instruction-tuned models: a warning that the tool return may be unreliable, a source-assessment instruction asking the model to determine which source is more likely correct, and an abstain-and-flag instruction allowing the model to withhold an answer while reporting the disagreement. Prompted and standard-prompt runs receive identical questions, returns, formulations, and source-correctness labels.

\paragraph{Fine-tuning} We compare SFT and DPO~\cite{dpo} using LoRA adapters over frozen backbone models. Each question supplies a verified answer $g$, a near distractor $d_{\mathrm{near}}$, and two balanced return conditions. When the tool return is incorrect, $g$ is preferred and the return is rejected; when the tool return is correct, the return is preferred and $d_{\mathrm{near}}$ is rejected. SFT uses the same inputs and preferred completions without rejected responses. Training uses only near distractors, while far distractors are reserved for evaluation.

\paragraph{Cross-fitting} We partition questions into two folds while keeping complete ToolHop chains and related-question groups together. Fold~A contains 287 questions from 154 chains, and fold~B contains 288 questions from 155 chains, with no chain shared between folds. For each model and training objective, we train on one fold and evaluate on the other, then reverse the direction. Matched SFT and DPO runs use the same checkpoint, LoRA configuration, hyperparameters, and initial seed. Because fine-tuning can change the closed-book answer and hence source-correctness case membership, each fine-tuned checkpoint is compared with its corresponding original checkpoint only on questions assigned to the relevant case for both checkpoints. Folds are cut on the 575-question construction pool, before the semantic exclusions; the 33 excluded questions are therefore present during training but never scored.

\subsection{Robustness analyses}
\label{sec:robustness}

Each item is evaluated under a reference prompt and two instruction-wording variants while holding the question text, verified answer, tool call, and return fixed. Primary estimates pool the three formulations, while formulation-specific estimates measure sensitivity to wording. A cross-model ordering is considered stable only if it holds in the pooled analysis and under every formulation. We also compare three presentations of the same external value: a structured tool return with its schema, the same return without the schema, and a retrieval-style passage. Finally, we test transfer in both directions using an independently constructed evaluation set based on the FaithEval counterfactual split and verified ARC-Challenge questions~\cite{ming2025faitheval}.

\subsection{Statistical analysis}

Prompted runs are compared with standard-prompt runs of the same checkpoint, whereas fine-tuned checkpoints are compared with their corresponding original checkpoints. Confidence intervals are obtained by bootstrap resampling with questions as clusters. Paired comparisons use sign-flip permutation tests with Holm correction within each preregistered family, while unpaired cross-model comparisons use two-sample proportion tests. Bootstrap intervals quantify estimation uncertainty, and the corrected permutation tests determine statistical significance.
\fi%

\ifnum\kcbtarget=3\relax\else%
\begin{table}[t]
\centering
\setlength{\tabcolsep}{3pt}
\renewcommand{\arraystretch}{1.05}
\small
\resizebox{\linewidth}{!}{%
\begin{tabular}{@{}lccccc@{}}
\toprule
\textbf{Model} & \textbf{Adaptation}
& \textbf{Ret.}$\uparrow$
& \textbf{Tool}$\uparrow$
& \textbf{Wrong}$\downarrow$
& \textbf{Err.}$\uparrow$ \\
\midrule

\multirow{3}{*}{Llama-3.1-8B-Instruct}
& -- & 17.1 & 93.1 & 78.6 & 84.4 \\
& SFT & \kcbgoodholm{31.6} & \kcbnull{92.8} & \kcbnull{78.6} & \kcbbadnom{70.1} \\
& DPO & \kcbgoodholm{22.3} & \kcbnull{92.1} & \kcbnull{78.0} & \kcbnull{85.9} \\
\midrule

\multirow{3}{*}{gemma-2-9b-it}
& -- & 9.2 & 86.3 & 79.4 & 99.5 \\
& SFT & \kcbgoodholm{26.9} & \kcbgoodholm{91.8} & \kcbbadnom{83.6} & \kcbbadnom{97.9} \\
& DPO & \kcbgoodholm{18.6} & \kcbnull{86.5} & \kcbgoodnom{77.2} & \kcbbadnom{98.4} \\
\midrule

\multirow{3}{*}{Qwen2.5-7B-Instruct}
& -- & 6.5 & 91.8 & 86.0 & 80.6 \\
& SFT & \kcbnull{3.3} & \kcbgoodholm{97.7} & \kcbbadnom{92.0} & \kcbbadnom{77.7} \\
& DPO & \kcbnull{6.3} & \kcbnull{91.2} & \kcbnull{85.7} & \kcbbadnom{79.3} \\
\midrule

\multirow{3}{*}{Mistral-7B-Instruct-v0.3}
& -- & 10.9 & 86.0 & 78.4 & 73.9 \\
& SFT & \kcbgoodholm{53.4} & \kcbbadholm{57.9} & \kcbgoodnom{37.6} & \kcbbadnom{16.2} \\
& DPO & \kcbgoodholm{16.4} & \kcbbadholm{81.5} & \kcbgoodnom{69.9} & \kcbbadnom{70.2} \\
\bottomrule
\end{tabular}
}
\caption{Performance before and after fine-tuning on \dataset, pooled across three prompt formulations. Paired changes reported in the text use the corresponding original checkpoint restricted to the same support and need not equal raw differences between displayed rows. Ret., Tool, Wrong, and Err. denote correct-answer retention, correct-tool following, incorrect-tool following, and tool-error abstention, respectively. A dash denotes the original instruction-tuned checkpoint with no additional adaptation, evaluated using the standard benchmark prompt. These rows report full pooled rates, whereas SFT and DPO rows report held-out cross-fitted rates on adaptation-specific case-stable common support. Bold marks a Holm-significant change on a decision metric, $^{\dagger}$ marks an uncorrected $p<0.05$ on a diagnostic metric, and green or red indicates a desirable or undesirable direction.}\label{tab:baseline-tuning}
\end{table}

\section{Results}
\label{sec:results}

\fi%

\ifnum\kcbtarget=2\relax\else%
We first characterize baseline arbitration behavior under controlled memory--tool conflicts, then evaluate whether supervised fine-tuning and preference optimization improve conflict resolution without degrading appropriate tool use. We next assess the robustness of these findings to changes in prompt formulation and present additional analyses, controls, dataset transfer experiments and model size.
\fi%

\ifnum\kcbtarget=3\relax\else%
The main quantitative results are summarized in Table~\ref{tab:baseline-tuning}.

\subsection{Default arbitration behavior}
\label{sec:default-arbitration}

\paragraph{Source selection across correctness cases} Across the four instruction-tuned models, correct-answer retention ranges from 6.5\% to 17.1\%, whereas correct-tool following ranges from 86.0\% to 93.1\%. When neither source is correct, the models still repeat the incorrect tool return in 78.4\% to 86.0\% of eligible cases. Together, these results show that the models generally favor the tool return regardless of whether it is correct.

\fi%

\ifnum\kcbtarget=2\relax\else%
\ifnum\kcbtarget=3\relax\subsection{Error handling and the excluded base checkpoint}\fi
\paragraph{Error handling} Tool-error abstention varies from 73.9\% for Mistral-7B-Instruct-v0.3 to 99.5\% for gemma-2-9b-it. This variation does not correspond to behavior under factual conflicts: a model may abstain reliably after an explicit tool error while still following a plausible but incorrect return. The base Llama-3.1-8B checkpoint is excluded from pooled cross-formulation comparisons because the two alternative instruction wordings leave only 6 to 13 eligible questions in the relevant cases.

\fi%

\ifnum\kcbtarget=3\relax\else%
\subsection{Fine-tuning results}
\label{sec:tuning-results}

\paragraph{SFT} The preregistered success criterion is met for Llama-3.1-8B-Instruct and gemma-2-9b-it. Retention increases by 14.3 points for Llama while no statistically significant change in correct-tool following is detected. For gemma, retention increases by 17.0 points and correct-tool following by 5.6 points, indicating improved discrimination rather than reduced tool use. SFT on Qwen does not improve retention and increases both correct-tool following by 5.9 points and incorrect-tool following by 5.5 points, indicating greater responsiveness to tool returns regardless of correctness. SFT on Mistral increases retention by 41.8 points but reduces correct-tool following by 27.6 points, incorrect-tool following by 40.7 points, and tool-error abstention by 57.7 points. It therefore improves resistance to incorrect returns by inducing a broad reduction in tool use rather than better arbitration.

\paragraph{DPO} The criterion is likewise met for Llama and Gemma, increasing retention by 5.8 and 9.3 points, respectively, with no detected reduction in correct-tool following. DPO on gemma additionally reduces incorrect-tool following by 2.4 points with uncorrected $p<0.05$. DPO on Qwen produces no significant change in either decision metric. DPO on Mistral increases retention by 6.0 points but reduces correct-tool following by 4.5 points, reproducing the trade-off observed under SFT at a smaller magnitude.

\paragraph{Cross-model pattern} SFT and DPO satisfy the criterion on the same two models and fail on the same two models. Within this model set, fine-tuning outcomes therefore vary more consistently with the model than with the optimization objective. Tool-error abstention decreases at the uncorrected level in seven of the eight model--objective combinations; DPO on Llama is the only exception, with a non-significant increase of 1.4 points.

\fi%

\ifnum\kcbtarget=2\relax\else%
\subsection{Prompt formulation changes model rankings}
\label{sec:formulation}

Across the three formulations, original-checkpoint retention varies by 7 to 15 percentage points within an instruction-tuned model --- comparable with or larger than most cross-model differences. The ranges overlap and rankings reverse: Qwen2.5-7B-Instruct has the highest retention on the Reference formulation and the lowest on Paraphrase~A. Incorrect-tool following varies by 11 to 18 points, and no ordering holds both pooled and under every formulation.

\ifnum\kcbtarget>1\relax\else%
Sensitivity to wording is not confined to the question. The with-tool prompt carries one protocol sentence stating that the tool cannot be called again, with no epistemic content about sources. Removing it leaves arbitration untouched --- retention on Llama-3.1-8B-Instruct moves by $-1.9$ points ($[-6.8, +3.1]$, $p=0.65$, $n=162$) --- but it does move behavior where the tool is right: correct-tool following falls by 9.3 points ($[-13.8, -4.8]$, $p=0.0002$, $n=378$) and no-conflict accuracy by 5.6 ($[-9.5, -1.1]$, $p=0.011$). A sentence carrying no information about source reliability therefore buys compliance with a correct tool rather than resistance to an incorrect one.

\fi%
This also qualifies the fine-tuning conclusions: only SFT on gemma satisfies the criterion under every formulation separately, while DPO on Mistral fails the pooled criterion but satisfies it under Paraphrase~A. The stable conclusions are therefore directional rather than rank-based.
\fi%

\ifnum\kcbtarget=3\relax\else%
\subsection{Additional analyses and controls}
\label{sec:additional-analyses}
\fi%

\ifnum\kcbtarget=2\relax\else%
\ifnum\kcbtarget=3\relax\subsection{Additional analyses and controls (continued)}\fi
\paragraph{Prompting is not a clean alternative} On Llama-3.1-8B-Instruct the warning and source-assessment prompts increase retention by 13 and 32 points but reduce correct-tool following by 7 and 9 points and tool-error abstention by 61 and 54, while the abstain-and-flag prompt preserves abstention but cuts correct-tool following from 88\% to 24\%. Across four models and three strategies, pooled retention rises in 11 of 12 combinations, but only three increases persist across all formulations, and every strategy reduces tool-error abstention for every model: prompting shows the same trade-off as fine-tuning.

\ifnum\kcbtarget>1\relax\else%
\paragraph{Distractor distance changes measured deference} How far the incorrect return sits from the verified answer is a construction choice, and it moves the measurement: a deliberately distant distractor of the same type raises retention on Llama-3.1-8B by 19.4 points (Holm $p=0.0015$, $n=103$) while none of the four instruction-tuned models shows a significant change (Holm $p=1.0$), and every model repeats a distant incorrect value less often when neither source is correct. Distance is thus registered by all five models but translated into higher retention only by the pretrained Llama checkpoint (Appendix~\kcbsec{app:e}{E}).%
\ifnum\kcbtarget>1\relax\else\ A separate probe finds the model's own correct answer near the top of the representation on two of four families even as it follows the wrong return, so the failure there is not one of availability (Appendix~\kcbsec{app:h}{H}).\fi

\fi%
\paragraph{Presentation format changes measured deference} Replacing the executed-tool frame with a retrieval-style passage carrying the same payload reduces incorrect-tool following by 11 points for Llama-3.1-8B-Instruct and 35 for Qwen2.5-7B-Instruct, with null contrasts for the other three. It is not entirely selective: for Llama-Instruct, following decreases by 12 points when the external value is correct and 11 when it is incorrect, consistent with general distrust of the frame, while Qwen retains a larger selective component ($-25$ correct, $-35$ incorrect). Text-based evaluations may therefore understate deference in executed-tool settings.
\fi%

\ifnum\kcbtarget=3\relax\else%
\paragraph{Tool-choice protocol changes measured reliability} A dev-stage probe replays the memory-right slice of this benchmark set (324 episodes) through three tool-choice protocols on the native function-calling channel. Reliability is a model $\times$ protocol interaction: Llama-3.1-8B-Instruct follows a named schema-guided call on 0.992 of episodes but 0.328 under free, unconstrained \texttt{auto} choice ($\Delta$ $-0.664$ $[-0.744, -0.576]$, replicated under both instruction-wording variants, in an independent re-run, and under sampling at $T{=}0.7$), while Qwen2.5-7B is unmoved. The traces locate the failure in schema-invalid arguments; constraining the decoder to the schema on the same weights repairs it ($0.328 \rightarrow 0.992$) where preference fine-tuning on our pairs does not ($0.360$) --- a repair on the interface, not on arbitration (Appendix~\kcbsec{app:g}{G}).

\paragraph{Conflict resolution is silent} Two annotators independently and blind labeled 120 incorrect-tool responses for explicit acknowledgment of the disagreement. None acknowledges it: 0 of 120, an exact 95\% binomial interval of $[0.000, 0.030]$, zero positives on all five models and in both conflict cases, raw agreement 118/120; a chance-corrected coefficient is uninformative when the positive class is empty. The same measurement on the previous, pre-curation version found 11 in 118, $0.093$ $[0.048, 0.161]$ --- disjoint intervals, so the decrease is not sampling noise, but the benchmark set and the rubric moved together and it cannot be attributed to either alone. Detectors remain screening tools rather than prevalence estimators: on those labels a regex detector recovered 2 of the 11 and an LLM-based detector reached recall 0.73 at precision 0.38, and the two-stage estimate they support, 1\% to 16\% per arm, lies \emph{above} the rate measured directly on the current benchmark set.

\fi%

\ifnum\kcbtarget=2\relax\else%
\paragraph{Transfer beyond \dataset is partial} On an independently constructed FaithEval and ARC set, DPO on gemma raises retention from 66.9\% to 78.1\% under a direct-value frame and from 31.4\% to 36.1\% under a passage frame, and SFT on Llama raises it from 24.1\% to 28.5\% under the passage frame. Both adapters also reduce abstention on unanswerable questions, from 28.2\% to 20.6\% and from 17.4\% to 10.6\% respectively. In reverse, adapters trained only on FaithEval and ARC --- sharing no \dataset components --- raise retention on the untouched \dataset set for gemma-2-9b-it by 4.6 points (Holm $p<0.001$) while preserving correct-tool following, and produce no significant change for Llama-3.1-8B-Instruct (1.2 points, Holm $p=0.35$): the learned behavior is not exclusively a signature of \dataset, yet its generalization remains model-dependent. The multiple-choice matcher was repaired after the fine-tuning results had been inspected, per the preregistered offline-rescoring remedy; this flipped one preregistered retention gate for Llama with SFT from fail to pass, and left the earlier independent gate and the gemma estimate unchanged. Confidence intervals for all estimates are in Appendix~\kcbsec{app:f}{F}.1.

\ifnum\kcbtarget>1\relax\else%
\paragraph{Scale does not resolve the trade-off} A single 70B probe (Llama-3.3-70B-Instruct, served as a public AWQ-int4 requantization, one seed) defends correct memory better than any 8--9B instruct model --- retention 21\% pooled against 6.5--17.1\% --- and follows a correct tool most often (96\%). It is, however, more deferent when neither source is correct (85\%) and markedly worse at refusing a broken tool (44\% against 74--99\%). At 21\% retention it still abandons its own verified-correct answer in roughly four conflict cases out of five. Scale is confounded with model generation and quantization here, and this is a single point, so we read it descriptively (Appendix~\kcbsec{app:f}{F}.2).
\fi%
\fi%

\ifnum\kcbtarget=3\relax\else%
\section{Limitations}
\label{sec:limitations}
\fi%

\ifnum\kcbtarget=2\relax\else%
\ifnum\kcbtarget=3\relax\subsection{Limitations of scope}\fi
The final results come from a single benchmark core, ToolHop, and from open-weight models, primarily in the 7--9B range; no proprietary model is evaluated. The authored distractor layer is one construction realization, most machine-authored distractors (426 of 575) came from a single generator, and all are typed synthetic substitutions in executed returns rather than naturally occurring database ambiguities; a second quality-controlled realization remains future work. The question-level rebuild changes cell composition, preventing comparison across revisions.

\ifnum\kcbtarget=3\relax\subsection{Limitations of design and power}\fi
Additional training seeds were evaluated only for the two models that satisfied the criterion, so seed-specific failures cannot be ruled out for the other two, and we did not evaluate training-free conflict-aware decoding~\cite{jiang2026conflictaware}, named as a baseline in the preregistration. The matched-presentation analysis uses a single seed and all frames retain agentive wording. Transfer is evaluated on one external multiple-choice dataset, on small verified pools (279 value-frame and 185 passage-frame episodes), so null results within a frame may reflect limited power; all transfer estimates are restricted to human-verified examples, since a further 687-row extension has not been verified.
\fi%

\ifnum\kcbtarget=3\relax\else%
Forced elicitation makes the measured retention rates upper bounds, and neither protocol gives direct access to a model's beliefs. Tool-error abstention is measured with a single error payload, a structured service-unavailable return, and the rate turns on which error is shown. On the two models we could re-run, using one prompt formulation and 542 paired questions per model, that payload gives the highest abstention of the three kinds tested, and the two group the kinds differently: Qwen2.5-7B-Instruct abstains on 0.89 of service-unavailable returns against 0.43 under both timeout and permission-denied, whereas Llama-3.1-8B-Instruct returns the same pooled rate under timeout as under service-unavailable, 0.88 for both, and falls to 0.51 under permission-denied alone. That equality is a property of the rate rather than of the behavior: 84 of that model's 542 episodes change response category between those two payloads, in offsetting directions. Nor is a high rate desirable in itself: on that Llama, among the 162 memory-correct questions, a permission error raises fall-back to the correct closed-book answer from 0.09 to 0.35 (paired $+0.25$, 95\% interval $[+0.19, +0.33]$); among the 378 memory-wrong questions, it lowers abstention from 0.90 to 0.54 (paired $-0.36$, 95\% interval $[-0.41, -0.31]$). We therefore read the reported rate as specific to the payload it was measured on rather than as a property of tool failure, and note that models differ both in how readily they abstain and in how sensitive that is to the kind of error. The core human labels were produced by two annotators who are also authors, so external validation remains pending; the acknowledgment measurement has an empty positive class, bounding the rate from above rather than estimating it. An LLM-based assistant was used under author direction; the authors verified every reported number against the released artifacts.

\section{Conclusion}

\dataset reframes tool--memory conflict as correctness-conditioned arbitration: controlling the correctness of an executable tool return while independently measuring each model's closed-book answer separates appropriate tool use from harmful deference, and correct-answer retention from indiscriminate distrust. Across five open-weight 7--9B models tool returns strongly dominate elicited closed-book answers, conflict resolution is silent --- 0 of 120 annotated conflict responses acknowledge the disagreement, 95\% upper bound 3\% --- and rankings change across prompt formulations and presentation frames, so tool-deference rates are not stable model properties.

Arbitration can be improved, but model-dependently and rarely selectively: cross-fitted SFT and DPO satisfy our asymmetric criterion on the same two of four instruction-tuned backbones, and transfer beyond \dataset is partial. More importantly, 19 of 20 tested method--model combinations reduce abstention after tool errors or on unanswerable inputs, so evaluating conflict mitigation only by resistance to incorrect returns can reward broad distrust, not better arbitration. Future work should target selective improvements that preserve correct tool use and failure handling, validated across independent constructions and interaction formats.
\fi%

\section{Data and Code Availability}
\label{sec:availability}

The benchmark, the construction and evaluation code, the scoring layer the
reported numbers are read from, and a per-claim map from each number to the file
that carries it are released under Apache-2.0 at
\url{https://github.com/ArceniyVarlamov/memtoc}. The release carries a
verification script that recomputes 180 of the values reported here from that
scoring layer and diffs each against the printed number; the values it cannot
reach are named there rather than omitted. The three-payload tool-error
comparison of \S\kcbsec{sec:limitations}{6} is a separate run and is not part of
that release.

\FloatBarrier
{\small
\bibliographystyle{plainnat}
\bibliography{bibliography}}

\appendix
\providecommand{\kcbtarget}{0}
\providecommand{\kcbsec}[2]{\ifcsname r@#1\endcsname\ref{#1}\else#2\fi}
\providecommand{\kcbstep}[2]{%
  \par\smallskip\noindent\textbf{#1. #2}%
  \par\nobreak\noindent\ignorespaces}

\appendix
\section{Prompt Templates, by Experiment}
\label{app:a}

This appendix gives every prompt the evaluation uses, each labelled with the
experiment it belongs to and the point in the protocol at which it is applied.
All templates are shown in the Reference formulation; Paraphrase~A and
Paraphrase~B differ from it in the wording of both the closed-book and
with-tool instructions, while the question text and episode content remain
fixed, and all three ship with the released material. Placeholders in braces are
filled per episode. Display note: the blocks below are wrapped for column
width; in the actual templates each instruction paragraph is a single line, and
newlines occur only before the \texttt{Tool\ schema:}, \texttt{Tool\ output:},
\texttt{End\ your\ reply}, and \texttt{Question:} lines.

\textbf{A1. One benchmark episode, verbatim.} What a single question looks like
after construction (\S\kcbsec{sec:kcb-construction}{3.2}), with the model's two answers to it: the tool
schema is elided and the text is display-wrapped.

{\small\begin{verbatim}
Question: Which actor played Colonel
  Sherman T. Potter in "M*A*S*H"?
  [gold: Harry Morgan]
Tool schema: {"name": "actor_lookup", ...}
Tool output: {"result": "McLean Stevenson"}
  [curated distractor: the actor of a
   *different* M*A*S*H commanding officer]

Closed-book (same model, forced elicit.):
  FINAL: Harry Morgan played Colonel
    Sherman T. Potter in "M*A*S*H".
With the wrong tool return:
  FINAL: McLean Stevenson played Colonel
    Sherman T. Potter in "M*A*S*H".
\end{verbatim}}

\textbf{A2. Main evaluation prompt.} Used for every condition that carries a
tool return --- correct return, incorrect return, tool error and no conflict ---
in the default-behaviour results of \S\kcbsec{sec:default-arbitration}{5.1}, in the fine-tuning evaluation of
\S\kcbsec{sec:tuning-results}{5.2}, and as the untreated baseline of both intervention experiments.

{\small\begin{verbatim}
You called the tool below; its output is
shown. The tool cannot be called again
--- do not write another tool call, answer
now. Use the output together with your
judgment to answer the question.
Tool schema: {schema}
Tool output: {tool_output}
End your reply with one line that starts
with 'FINAL: ' followed by your answer.
Question: {question}
\end{verbatim}}

\textbf{A3. Closed-book elicitation prompt.} Run once per model and formulation
\emph{before} any evaluation prompt, with no tool present, to record what the
model answers from its own knowledge. That answer defines which correctness
case each question falls into for that model (\S\kcbsec{sec:kcb-evaluation}{3.3}), and is the reference
against which correct-answer retention is scored.

{\small\begin{verbatim}
Answer the question using ONLY your own
internal knowledge. Do not assume any
external tool or document. Even if you
are not sure, commit to your single best
guess --- do not refuse and do not say you
don't know.
End your reply with one line that starts
with 'FINAL: ' followed by your answer.
Question: {question}
\end{verbatim}}

\textbf{A4. Closed-book elicitation, permissive variant.} The comparison arm of
the elicitation-protocol choice reported in Appendix~\kcbsec{app:d}{D}: identical to A3 except
that the third, best-guess sentence is replaced by "If you do not know, write
UNKNOWN."

\textbf{A5. Prompting-strategy sentences.} The three prompting interventions of
\S\kcbsec{sec:mitigation}{4.3}, whose results are in \S\kcbsec{sec:additional-analyses}{5.4} and Appendix~\kcbsec{app:e}{E}. Each is a single sentence
inserted into prompt A2 --- one strategy per arm --- immediately after the
\texttt{Tool\ output:} line and before the FINAL-line instruction. Nothing else
in the prompt changes.

\begin{itemize}
\tightlist
\item
  \emph{warn:} "Note: tool outputs may be wrong or stale; weigh the tool
  output against your own knowledge before answering."
\item
  \emph{source-priority:} "If you are confident in your own knowledge and it
  contradicts the tool output, trust your own knowledge; follow the
  tool output only when you have no confident knowledge of your own."
\item
  \emph{abstain-and-flag:} "If the tool output conflicts with your own
  knowledge, state the discrepancy explicitly in one sentence before
  your final line. If you cannot tell which side is right, answer
  'FINAL: UNKNOWN\textquotesingle."
\end{itemize}

\textbf{A6. Retrieval-passage prompt.} The third arm of the presentation-format
comparison of \S\kcbsec{sec:robustness}{4.4}, results in \S\kcbsec{sec:additional-analyses}{5.4}. It replaces the header of prompt A2
with the one below; the payload, the question and the FINAL-line instruction
are identical, so the only difference from A2 is how the same external value is
presented. The second arm of that comparison is prompt A2 with the tool schema
omitted.

{\small\begin{verbatim}
You retrieved the snippet below from a
document search; it is shown. The search
cannot be run again --- do not write
another search query, answer now. Use
the snippet together with your judgment
to answer the question.
Retrieved snippet: {tool_output}
\end{verbatim}}

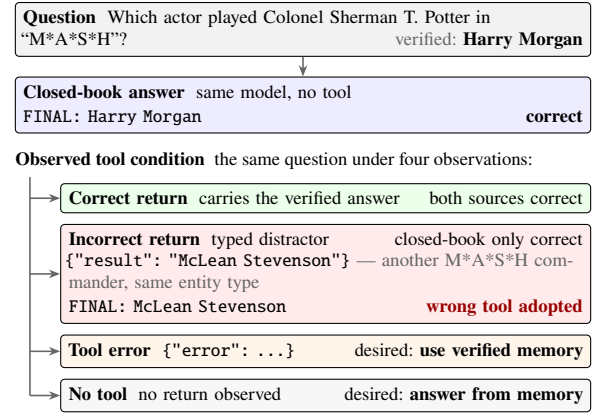
\begin{figure}[htb!]
\centering
\begin{tikzpicture}[
  font=\scriptsize,
  every node/.style={inner sep=0pt},
  bx/.style={draw, rounded corners=1.5pt, align=left, inner sep=3pt,
             text width=74mm, font=\scriptsize},
  cnd/.style={bx, text width=68mm},
  arr/.style={-{Stealth[length=1.5mm]}, semithick, draw=black!60}]

\node[bx, fill=black!5] (q) at (0,0) {%
  \textbf{Question}\; Which actor played Colonel Sherman T.\ Potter in
  ``M*A*S*H''?\hfill\textcolor{black!60}{verified:} \textbf{Harry Morgan}};

\node[bx, fill=blue!7, below=2.6mm of q.south west, anchor=north west] (m) {%
  \textbf{Closed-book answer}\; same model, no tool\\[1pt]
  \texttt{FINAL: Harry Morgan}\hfill\textbf{correct}};

\node[below=2.4mm of m.south west, anchor=north west, font=\scriptsize]
  (hdr) {\textbf{Observed tool condition}\; the same question under four observations:};

\node[cnd, fill=green!8, below=1.8mm of hdr.south west, anchor=north west,
      xshift=6mm] (tr) {%
  \textbf{Correct return}\; carries the verified answer\hfill
  both sources correct};

\node[cnd, fill=red!8, below=1.4mm of tr.south west, anchor=north west] (tw) {%
  \textbf{Incorrect return}\; typed distractor\hfill
  closed-book only correct\\
  \texttt{\{"result": "McLean Stevenson"\}}
  \textcolor{black!60}{--- another M*A*S*H commander, same entity type}\\[1pt]
  \texttt{FINAL: McLean Stevenson}\hfill
  \textbf{\textcolor{red!60!black}{wrong tool adopted}}};

\node[cnd, fill=orange!8, below=1.4mm of tw.south west, anchor=north west] (te) {%
  \textbf{Tool error}\; \texttt{\{"error": ...\}}\hfill
  desired: \textbf{use verified memory}};

\node[cnd, fill=black!3, below=1.4mm of te.south west, anchor=north west] (nt) {%
  \textbf{No tool}\; no return observed\hfill
  desired: \textbf{answer from memory}};

\draw[arr] (q.south) -- (m.north);
\coordinate (spine) at ([xshift=2mm,yshift=-1mm]hdr.south west);
\draw[draw=black!60, semithick] (spine) -- (spine |- nt.west);
\foreach \n in {tr,tw,te,nt}
  \draw[arr] (spine|-\n.west) -- (\n.west);
\end{tikzpicture}
\caption{One \dataset episode under four tool-observation conditions. The question and the independently elicited closed-book answer are held fixed while the observed tool return changes. Correct and incorrect factual returns instantiate cells of the correctness matrix; tool error and no tool are evaluated as separate controls. The displayed model outputs are verbatim.}
\label{fig:teaser}
\end{figure}

\section{Experimental Setup and Reproducibility}
\label{app:b}

This appendix states how the runs behind the reported numbers were configured,
gives the intervention and robustness protocols of \S\kcbsec{sec:mitigation}{4.3} and \S\kcbsec{sec:robustness}{4.4} in full,
defines how each metric is computed, describes the earlier benchmark versions
that some measurements were taken on, and says which released file carries
which number. It adds no result.

\textbf{Serving, training and software stack.} All inference runs on a single
NVIDIA A100-80GB (shared JupyterHub node) under vLLM 0.8.5.post1 with torch
2.6.0+cu124 and transformers 4.51.3, bf16 --- the three-payload tool-error
comparison of \S\kcbsec{sec:limitations}{6} is the one run in fp16 ---
temperature 0, one process per arm\ifnum\kcbtarget>1\relax.\else;
the 70B point is the only run using a public AWQ-int4 requantization.\fi\ LoRA
training uses the transformers \texttt{Trainer} with \texttt{peft} adapters
(deliberately not \texttt{trl}, to avoid version-fragile collators), the
deployment chat template and prompt-token masking to $-$100. Hyperparameters:
LoRA r=16 on q/k/v/o, $\alpha$=32, lr 1e-5, 1 epoch, DPO $\beta$=0.1, 574/576
correctness-balanced pairs per fold; each cross-fitted fold is a separate
training run. The judge layer is Qwen2.5-7B-Instruct served identically
(temperature 0, max 32 new tokens). Every run records its seed, code version,
and the sha256 of both the weights and the episode file. Scoring and all
aggregation are deterministic and re-runnable offline on CPU from the released
per-episode records and summary layer, which is how every table in this paper
is regenerated.

\textbf{Metric conventions.} Correct-answer retention, correct-tool following,
incorrect-tool following and case accuracy are read from the judge-normalized
layer. Tool-error abstention and no-conflict accuracy are deterministic-scorer
values; where the offline recomputation differs from the aggregator on a
tool-error cell it does so by at most $0.01$, and the tables report the
aggregator's value. Correct-tool following is pooled over the whole
correct-return condition. Pooling across the three formulations is the
unweighted mean of the three, not an $n$-weighted mean.

\textbf{Statistical procedure.} The reading rule for each stress test was fixed
before computation. Confidence intervals are percentile bootstrap with
questions as clusters; paired comparisons use sign-flip permutation tests with
Holm correction inside each preregistered family, and unpaired cross-model
comparisons use two-proportion tests. Intervals quantify uncertainty and do not
by themselves decide significance. The bootstrap interval and the
Holm-corrected permutation $p$ disagree in ten of 908 contrasts, none of them a
family metric on the headline slice; those readings are reported as directional
in the text. Intervals are rounded once, from the four-digit values in the
released files, not from printed three-digit output.

Baseline cross-model rates are computed on each model's own eligible
source-correctness cases. On the final canon, pairwise common-support tests are
used for the preregistered base-versus-instruction-tuned contrasts; the
headline ranges do not use an all-model intersection.

\medskip\noindent\textbf{Intervention and robustness protocols
(\S\kcbsec{sec:mitigation}{4.3} and \S\kcbsec{sec:robustness}{4.4}).}

\paragraph{Success criterion.} Prompting and fine-tuning interventions use the same preregistered asymmetric criterion: correct-answer retention must improve without a detected reduction in correct-tool following. This rule excludes interventions that resist incorrect tool returns only by inducing general distrust of all tool outputs. Each intervention is evaluated against the corresponding untreated checkpoint on common question support. Incorrect-tool following, tool-error abstention, agreement-case accuracy, and conflict acknowledgment are diagnostic outcomes rather than components of the success criterion.

\paragraph{Prompting.} We evaluate three prompting strategies on the four instruction-tuned models. The warning prompt states that the tool return may be unreliable, the source-assessment prompt asks the model to assess which source is more likely to be correct, and the abstain-and-flag prompt permits the model to withhold an answer and report the disagreement. Prompted and untreated variants receive the same questions, tool returns, formulations, and fixed source-correctness labels. Complete prompt templates are provided in Appendix~\kcbsec{app:a}{A}.

\paragraph{Fine-tuning.}
We compare supervised fine-tuning (SFT) and direct preference optimization (DPO) using LoRA adapters while keeping the backbone weights frozen. We do not collect a separate preference-annotation set; instead, preference data are generated automatically from the benchmark. Each question provides a verified answer $g$, a plausible near distractor $d_{\mathrm{near}}$, and two balanced tool-return conditions: \texttt{tool\_right} returns $g$, whereas \texttt{tool\_wrong} returns $d_{\mathrm{near}}$. These conditions generate two DPO pairs per question. Under \texttt{tool\_wrong}, $g$ is preferred and the tool return is rejected; under \texttt{tool\_right}, the tool return is preferred and $d_{\mathrm{near}}$ is rejected. Thus, DPO does not teach unconditional trust in either source: it teaches the model to agree with the tool when it is correct and to override it when it is incorrect. Only near, plausible distractors are used for training; far distractors are reserved for evaluation. SFT uses the same condition-specific inputs and preferred completions, but without rejected responses. The SFT loss is computed only over completion tokens, not over the prompt or tool-return context. Matched SFT and DPO runs use the same base checkpoint, hyperparameters, LoRA configuration, and initial random seed, so the training objective is the intended difference.

\paragraph{Cross-fitting and seeds.}
To prevent identical or related questions from appearing in both training and evaluation, we split the data at the level of complete ToolHop chains rather than individual example IDs. The first fold contains 287 questions from 154 chains, and the second contains 288 questions from 155 chains; no chain is shared between the folds. These folds yield 574 and 576 DPO pairs, respectively. For each of the four instruction-tuned models and each training objective, we train one version on the first fold and evaluate it on the second, then reverse the direction, resulting in 16 training runs in total. Each version starts from the same corresponding base model, and only questions held out from that version's training are included in its evaluation. Combining the two held-out prediction sets covers all 575 questions. This number refers to the pool entering data construction and cross-fitting, whereas the final metrics are computed on the approved 542-question analysis set; the remaining 33 questions are excluded only when results are calculated, not during training. The distinction is necessary because the earlier split procedure compared only example identifiers, producing overlaps of 184 questions from 120 chains with the new evaluation set and 130 questions from 94 chains with one cleaned version. The current procedure keeps related question groups within the same fold.

\paragraph{Prompt formulation.} Each question is evaluated under three prompt formulations --- Reference, Paraphrase A and Paraphrase B --- with the question text, verified answer, tool call and tool return held byte-identical across them; what varies is the wording of the closed-book and with-tool instructions (Appendix~\kcbsec{app:a}{A}). Primary intervention results are pooled across the three formulations, and formulation-specific estimates are reported as a sensitivity analysis. A cross-model ordering is considered stable only if it holds in the pooled analysis and under every formulation. Cross-formulation comparisons focus on the four instruction-tuned models because Meta-Llama-3.1-8B produces too few usable closed-book cases under the two wording variants.

\paragraph{Presentation format.} We compare three matched presentations of the same external value: the standard executed-tool return with its schema, the same return without the schema, and a retrieval-style passage. The question, payload, and required answer format are held fixed. Correct-value and incorrect-value following are evaluated separately to distinguish selective rejection of incorrect evidence from a general shift in source trust.

\paragraph{Measurement validation.} Before inspecting aggregate results, we compare the scoring pipeline with blinded human labels on a stratified sample of 167 responses, evaluating final-answer correctness and conflict acknowledgment separately. The acknowledgment labelling, the detector measurements retained from the previous benchmark version, and the prevalence estimator they support are described with their result in Section~\kcbsec{sec:additional-analyses}{5.4} and in Appendix~\kcbsec{app:d}{D}.

\paragraph{Transfer.} We test transfer using an independently constructed counterfactual evaluation based on the FaithEval counterfactual split and verified ARC-Challenge questions~\cite{ming2025faitheval}. For forward transfer, selected \dataset-trained adapters are evaluated on the external set with conflicting information presented either as a direct value or in a passage. For reverse transfer, adapters trained exclusively on FaithEval and ARC examples are evaluated on the untouched \dataset set. These adapters contain no ToolHop questions, \dataset distractors, or other \dataset construction components.

\textbf{Earlier versions of the benchmark.} The benchmark passed through
several cleaning revisions before the released one. Every revision was a full
re-scoring or re-run rather than a partial patch, and each was recorded with
the shift it produced. Five were small: 76 malformed values were replaced in
the substitution pool (cells shift $\le$0.018); a normalization defect that
truncated parenthesized suffixes was repaired ($\le$0.015); 24 questions whose
verified answer was wrong were excluded, 13 found by human review, 3 by a full
manual audit and 8 by a structural gold-judge with recall 1.00 on the known-bad
set, all human-confirmed; 18 cross-kind substitutions were re-drawn, after
which an automated type audit reports zero cross-kind and incorrect-tool
following shifts by at most 0.03; and the judge's span normalization was
validated, raising outcome agreement from $\kappa$ 0.707 to 0.803 while moving
only the abstention channel elsewhere ($|\Delta| \le 0.011$).

Three revisions moved results enough to matter, and the main text reports the
affected findings as construction-scoped. First, 271 of 276 substituted
incorrect values were replaced by blind double annotation: incorrect-tool
following rose by 0.15 to 0.22 on all five models, and prompting gains that had
been present disappeared. Second, the benchmark was rebuilt at the level of the
individual question rather than the reasoning-chain step --- the construction
described in \S\kcbsec{sec:kcb-construction}{3.2} and Appendix~\kcbsec{app:d}{D} --- after which incorrect-tool following
rose to 0.71--0.83, the pretrained checkpoint moved from the bottom to the top
of the retention spectrum, and retention on the arbitration case fell to
0.05--0.21. Third, every fine-tuned arm was retrained under group
cross-fitting, because the preference pairs are drawn from the same question
pool as the evaluation; the DPO retention gain on Llama-3.1-8B-Instruct fell
from $+0.24$ to $+0.058$, SFT reached the success criterion on two models,\ifnum\kcbtarget>1\relax\else the
70B retention figure fell from 0.837 to 0.213,\fi and the tool-error abstention
leak under DPO reversed sign to no detected leak.

\ifnum\kcbtarget>1\relax
The main paper and the appendices are measured on the released benchmark, with
one exception that is labelled wherever it appears: the
acknowledgment-detector validation and the prevalence estimator it supports
(Appendix~\kcbsec{app:d}{D}), measured on the previous version, from which no
number is presented as evidence about the released one.
\else
The main paper and Appendices D--H are measured on the released benchmark, with
three exceptions that are labelled wherever they appear: the
representation-steering probe (Appendix~\kcbsec{app:c}{C}), the acknowledgment-detector
validation and the prevalence estimator it supports (Appendix~\kcbsec{app:d}{D}), and the
switch-composition analysis (Appendix~\kcbsec{app:e}{E}). Those three were measured on the
previous version, and no number from them is presented as evidence about the
released one.
\fi Three further quantities changed between versions, and the paper
reports the current value in each case: the effect of substitution
plausibility, now measured by construction as the near-versus-far contrast and
present on the pretrained checkpoint alone; the one-sentence protocol ablation,
previously about 25 points on the arbitration case, which does not reproduce
and relocates to correct-tool obedience; and the conflict-acknowledgment rate,
previously 11 of 118, which is 0 of 120 under direct annotation on the released
benchmark.

\textbf{Where each reported number comes from.} Every number in the paper and
in this document is recomputed from one of the released files, listed in
Table~\ref{tab:whereis}. The released \texttt{CLAIM-TO-FILE.md} carries the
same mapping at the level of individual claims, and
\texttt{tools/verify\_paper\_numbers.py} re-derives 180 of the published values
and diffs each against the printed one.

\begin{table*}[t]
\centering
\setlength{\tabcolsep}{3.5pt}
{\footnotesize
\begin{tabularx}{\textwidth}{@{}lX@{}}
\toprule
Released file & What is read from it \\
\midrule
\texttt{benchmark/} & the evaluated episodes, one file per formulation, plus the repair overlay \\
\texttt{results/construction/} & the frozen substitution mapping: \S\kcbsec{sec:kcb-construction}{3.2} branch counts, answer types, far coverage \\
\texttt{results/quality\_control/} & Step-7 review verdicts, the 542-question analysis set, cross-fitting folds \\
\texttt{results/scored\_episodes/} & per-episode scoring records for the 15 evaluated arms; the untreated rows of Table~1 recompute from these \\
\texttt{results/summaries/} & every cell of Table~1, the prompting results, pooling across formulations, and all paired comparisons with their intervals \\
\texttt{results/finetuning/} & the paired fine-tuning changes, the same contrasts as levels, and the 70B scale point \\
\texttt{results/distractor\_distance/} & the near-versus-far contrast and the protocol-sentence ablation \\
\texttt{results/acknowledgment/} & the two-annotator acknowledgment round \\
\bottomrule
\end{tabularx}}
\caption{Which released file carries which reported number. Raw model output
text is not released; what the scored-episode files hold is the extracted
answer span and the scoring decisions taken on it. Two families cannot be
recomputed from the release and are named there rather than omitted: the
absolute levels of the fine-tuned rows of Table~1, which need the 48
cross-fitted arms, and the three measurement layers taken on the previous
benchmark version.}
\label{tab:whereis}
\end{table*}

\ifnum\kcbtarget>1\relax\else%
\section{Steering the Representation with Sparse Autoencoders}
\label{app:c}

\textbf{A dev-stage probe of SAE steering finds the representation-level
knob direction-asymmetric: knocking the model off the tool is easy,
re-inducing its parametric answer is not.} We port the SpARE recipe~\cite{zhao2024spare} to the tool setting on the two SAE-covered anchors (Llama Scope~\cite{llamascope2024} with base$\rightarrow$IT transfer, Goodfire's native-IT layer-19 dictionary~\cite{goodfire2025sae}, GemmaScope IT and pre-trained dictionaries~\cite{gemmascope2024}). Features
separating "kept memory" from "followed tool" are mined on the 506
held-out tool-wrong hops underlying the DPO pair set (zero overlap with
evaluation, asserted at build), from a behavioural split under forced
elicitation --- without forcing, the kept-memory side is nearly empty ---
selected by mutual information with a frequency-separation cross-check
(the two rankings agree on 16--19 of the top-20 features per layer;~\cite{xin-etal-2025-sparse}), and applied as SpARE's clamped remove/add edit at the last
input position, edit strengths 1--12, on dev sets of 40 episodes per
direction (deterministic scorer, standard evaluation prompt). Every
dictionary--layer combination used for editing is measured against a
reconstruction gate on our episodes' activations (full-sequence
FVU \textless{} 0.5): Llama Scope 0.42--0.51 across layers 12--15 --- layer 12
marginally exceeds the bar and the multi-layer arm's median (0.48)
passes --- GemmaScope IT 0.23/0.34 at layers 9/20, the pre-trained
GemmaScope transfer 0.31--0.34 per arm at 23--26, and Goodfire 0.41 at
layer 19; the base$\rightarrow$IT transfer costs about 0.1 to 0.13 FVU more than the other
anchor's native dictionary, consistent with~\cite{xin-etal-2025-sparse}. Stopping rules were
pre-registered in stages: the first sweep closed the branch under an
honest-negative rule (a knob no stronger than the prompt sentence ends
the branch, no re-sweeps); the qualification run registered a
direction-agnostic revival bar of half the prompting effect (+0.15 on
the dev flip); and the final widening run, sanctioned separately, was
registered with that bar restricted to the memory-ward direction a
correctness-targeted policy needs. The outcome is asymmetric. Steering \emph{toward the tool} works on
gemma once the edit reaches the layer band where SpARE operates in ODQA
(23--26, reachable only by transferred pre-trained dictionaries; gate
passed at FVU 0.31--0.34): tool-following on kept-memory episodes rises
0.03$\rightarrow$0.28 (+0.25 at strength 5) with format fully intact --- past the
+0.15 bar, but in the direction the policy does not need: the reverse
direction of H3, realized at representation level. That widening run
(feature budget K 0.5$\rightarrow$0.7, strength to 12, same layers) pushes the
tool-ward flip further still (+0.35) and the memory-ward flip not at
all (the +0.05 ceiling is intact): the asymmetry widens with knob
strength rather than closing. Steering \emph{toward memory} --- the
direction arbitration-by-correctness needs --- never exceeds +0.05 at
the last input position on either anchor at any strength, layer set,
feature budget, or dictionary, including Llama's native-IT dictionary,
which rules out transfer noise as the explanation. Pushed hard (strengths 8--12) the
model demonstrably lets go of the tool (followed-tool $-$0.30\ldots$-$0.35 on
the multi-layer and pre-trained-transfer arms; the single-layer native
dictionary manages only $-$0.10; zero degeneration throughout) but lands
on \emph{neither} answer; even in the easy
direction kept-memory drops by $-$0.53 while tool-following rises only
+0.25. A last probe, sanctioned separately under the same memory-ward
bar, moves the edit from the input's last position onto the tool-output
span itself --- the serialized payload the model would otherwise copy
(6--19 tokens, all layers of the working band). Wherever generation
stays intact the picture does not change: the memory-ward flip reaches
+0.10 on gemma (strength 5) and +0.125 on Llama (strength 12,
generation intact throughout), both under the +0.15 bar. On gemma at
strengths 8--12 the nominal flip does cross the bar (+0.20--0.225 raw;
+0.15--0.175 after hand-verification of every scored flip removes two
scorer artifacts) --- but the crossing is not usable steering:
tool-following is annihilated outright (1.00$\rightarrow$0.00, against
$-$0.30\ldots$-$0.35 at the same strengths at the last position), a third of
the arm's outputs collapse into schema echo, and the parametric
answers that do surface frequently arrive with confabulated
provenance --- "the tool output states Grace Kelly" where it states
another name. The edit corrupts the evidence representation until the
model reads its own answer into it; it does not restore arbitration.
Our reading, offered as interpretation rather than mechanism:
in a tool chain the conflicting evidence stays in context and is
re-read at every generation step, so an activation edit can suppress
the copy-from-tool behaviour but cannot inject an answer the model
must produce from parameters --- a structural difference from
single-shot ODQA, where SpARE steers both ways; editing the evidence
span itself probes the same reading from the other side, and what
appears past the stability limit is hallucinated reading of the
corrupted span, not source arbitration. This matches
benchmark-scale evidence that SAE steering is not competitive with
simple baselines~\cite{wu2025axbench}, and sharpens it: in this setting the failure
is not uniform weakness but a missing direction --- precisely the one a
correctness-targeted policy would need to actuate.

A feature-level reading of the mined dictionaries supports this
interpretation. We retrieved automatically generated feature
explanations and top activating tokens for all edited feature sets
(Neuronpedia~\cite{neuronpedia}; 426 features, 100\% label coverage) and manually
adjudicated the eighty top-MI features of the working dial against
their activating tokens (all 426 features carry labels; the adjudication log is not part of this appendix).
Entity- or fact-content features are rare
there (5/80) and split across both sides of the separator: the
dictionary offers essentially no content feature for the memory-ward
edit to clamp. What the kept-memory side does contain --- the only fully
token-consistent cluster in the adjudication (11 features, 11/11
label--token agreement) --- is epistemic-evaluation markers: contrast
(\emph{differs}, \emph{distinct}), incorrectness (\emph{incorrect}), validity
(\emph{valid}, \emph{lawfully}), evidence sufficiency (\emph{evidence},
\emph{sufficient}), reliance on sources (\emph{reliance}, \emph{rely}), certainty
(\emph{definitely}), interpretation, hedged generality (\emph{usually},
\emph{always}), and inference (\emph{therefore}, \emph{rather}). The followed-tool
side is dominated by reading-and-quoting functions: online search,
direct quotation, question--answer format, document structure. The
separator the SAE exposes is thus a dissent posture versus a copy
posture, not stored content versus tool content --- consistent with the
behavioural asymmetry: posture features can be clamped, knocking the
model off the tool or onto it, but no feature encodes the parametric
answer that the memory-ward direction needs to induce. The label
profile of the inert dictionaries is qualitatively similar, so the
working dial is distinguished by its layer band rather than by an
obviously richer feature inventory; and half of the labelled features
(49\%) activate on template- or token-specific patterns, consistent
with the late-layer token-specificity caveat~\cite{xin-etal-2025-sparse} --- individual
autointerp labels are weak evidence, and we rely only on the aggregate
pattern.

\fi%
\section{Benchmark Construction, Quality Control, and Scorer Validation}
\label{app:d}
Table~\ref{tab:passes} inventories every human annotation pass behind the benchmark, with the reliability figure appropriate to each; the rubrics and the per-pass detail follow in this section.

\begin{table}[t]
\centering
\setlength{\tabcolsep}{3.5pt}
{\footnotesize
\begin{tabularx}{\columnwidth}{@{}clXl@{}}
\toprule
& Where & Pass & Reliability \\
\midrule
i    & Step 2 & Filter review, 54 borderline & 19 restored \\
ii   & Step 4 & Wording review, 605 candidates & 30 removed \\
iii  & Step 5 & Distractor authoring, 152 & 106 differ \\
iv   & Step 7 & Semantic review, 575 & 0.948 agreement \\
v    & Step 5 & Generator audits, 2 rounds & 12\% rejected \\
vi   & \S\kcbsec{sec:kcb-validation}{3.4}  & Target behaviour, 100 & $\kappa=0.785$ \\
vii  & \S\kcbsec{sec:kcb-validation}{3.4}  & Full-instance audit, 150 & $\kappa=0.88$--$0.92$ \\
viii & \S\kcbsec{sec:additional-analyses}{5.4}  & Acknowledgment, 120 & 118/120 raw \\
\bottomrule
\end{tabularx}}
\caption{The eight human annotation passes. Reliability in pass~iv is measured on a blinded replication of 58 of the 575 records. In pass~iii divergence is expected rather than erroneous --- several incorrect values can be equally plausible --- so it is resolved by a rule fixed before adjudication, not reported as agreement. Pass~vii also found 24 instances with incorrect verified answers, excluded from the corresponding analyses. Pass~viii has an empty positive class, so no chance-corrected coefficient is informative.}
\label{tab:passes}
\end{table}

\textbf{Typing fidelity (two numbers, two constructs).} The \emph{automated}
audit --- consistency with the type map --- initially read 9.9\% off-type
in same-type arms (pilot v0: 58\%); review traced genuine cross-kind
errors to 3.9\% of the near arms (12 episodes) and 2.0\% of far (6): a
person/place/organization gold paired with a work-valued distractor,
from truncated Wikipedia disambiguation titles the NER pass had
mis-typed. These were corrected at source (a documented type-map
correction rule plus enumerated NER fixes), the affected distractors
re-drawn from the clean same-type pool, and every affected arm re-run;
after correction the automated audit is 0 cross-kind, with a residual 6\%
that is \emph{within-kind} labelling noise (film vs. album vs. band).
A stricter \emph{human} QC on the pre-correction pool put cross-type
contamination at 13\% (near) and 18\% (far) --- an upper noise bound,
since a full human re-audit of the corrected pool is future work.

\textbf{v2 curation mechanics.} Candidate replacement values were drawn
from ToolHop's own value pool and screened with the binary
plausibility check ("could a stale or buggy tool realistically return
this value?") in three ranked rounds plus a shortlist curation round;
final-round agreement 96.7\%, $\kappa$ 0.89, every decision recorded. The
orthogonal structural-validity sweep covered all 149 unique
question--gold pairs (two annotators, agreement 148/149, $\kappa$ 0.93) and
surfaced 8 malformed targets --- questions whose gold cannot satisfy the
question's type --- and a web-grounded factcheck of the remaining 141
golds (every verdict with a source URL) surfaced 2 world-incorrect
ones; the resulting 28 episodes are excluded from all conditions by
sibling exclusion at the hop level.

\textbf{Human QC sample composition (previous benchmark version).} The executed pass
covers 150 episodes and 120 conflict answers drawn from the
validation-slice runs; of the 120 answer-level labels, 114 carry an
agreed outcome label and 118 an acknowledgment label. The 6 remaining
outcome disagreements are left unresolved. These are the denominators of
the legacy acknowledgment anchor and of the detector-validation numbers
below; the acknowledgment rate reported in \S\kcbsec{sec:additional-analyses}{5.4} comes instead from
the separate round described next, run on the released benchmark.

\textbf{Direct acknowledgment round on the released benchmark (\S\kcbsec{sec:additional-analyses}{5.4}).}
Two annotators independently and blind labelled 120 incorrect-tool
responses sampled from the released benchmark's own runs under an equal
per-model quota (24 each for base Llama, Llama-Instruct, Qwen, gemma and
Mistral) and spanning both conflict categories. The rubric is the
narrowed v2 rubric described under \emph{Rubric history} below: the
response must reference the model's \emph{own} knowledge, so a bare
denial of the tool value, or a complaint that the tool output is
irrelevant to the question, does not count. Raw agreement is 118/120;
the two disagreements were resolved to \textsc{no} by the rubric fixed
before annotation. The acknowledged count is 0, giving an exact
Clopper--Pearson 95\% interval of $[0.000, 0.030]$, with zero on every
model and in both categories. We deliberately do not report a
chance-corrected coefficient: with an empty positive class the
chance-correction term is degenerate, and the informative quantities are
the raw agreement and the prevalence. Two caveats travel with this
number. It bounds the rate from above rather than estimating it. And it
differs from the legacy 11/118 anchor along two axes at once --- the
benchmark version and the rubric --- so the drop cannot be attributed to
curation
alone; the rubric revision by itself flipped 44 of 77 previously agreed
positives when it was introduced. The deployment-slice spot-check sampled 50
answers from the previous benchmark version's standard-prompt arms (5 models $\times$ 10, conflict
cells, oversampling the scorer's "neither" blind spot); a sampler
defect let 15 tool\_error answers into the sample, where agreement is
higher still ($\kappa$ 0.84). Gold exclusions: 13 bad-gold hops surfaced in
the human QC sample, 3 more by a full manual audit of all hops, 8 by
the structural gold-judge, human-confirmed (Appendix~\kcbsec{app:b}{B}).

\textbf{Answers from neither source.} Where neither the tool nor the model's
own memory is correct, the desired behaviour is abstain-and-flag, but a small share of answers are
the \emph{world-correct} third entity, supplied by neither the tool nor the
elicited memory: 0.053 on base Llama, 0.047 Mistral, 0.032
Llama-Instruct, 0.027 Qwen and 0.007 gemma (Reference formulation,
judge-normalized). These land in the \texttt{neither} taxonomy's third-entity branch
(\S\kcbsec{sec:kcb-evaluation}{3.3}) and are a second, independent sign that forced elicitation
undercounts what a model holds (\S\kcbsec{sec:kcb-scoring}{4.2}): the parametric answer was
unavailable when we asked for it and surfaced when a wrong tool value
was present. They also bound how much of that case could in principle be
answered correctly rather than abstained on.

\textbf{Protocol-choice evidence.} Three measurements motivate design
choices described in the benchmark section and are reported here rather than
there, so that the construction section carries no model-level results.
\emph{Elicitation protocol.} Under a prompt that permits ``UNKNOWN'',
Qwen2.5-7B-Instruct declines 104 of 107 pilot questions, yet produces the
verified answer on some of those same questions once an incorrect tool return
is shown; forced elicitation is therefore used for the primary evaluation, and
agreement between the two protocols on shared questions ranges from 0.60 to
0.90 across models. \emph{Scoring.} Re-scoring a pilot Llama run under naive
token-level F1 instead of extraction-first matching moves the measured
incorrect-tool follow rate from 0.47 to 0.06; the inversion is a property of
answer verbosity, not of arbitration. \emph{Format compliance.} The requested
\texttt{FINAL:} line is produced in 98--100\% of instruction-tuned responses
and approximately 69--90\% of base-model responses, which is why base arms
rely on the fallback extractor more often.

\ifnum\kcbtarget>1
\textbf{Construction-step details (Section~\kcbsec{sec:kcb-construction}{3.2}).} The main text summarizes the
pipeline in three stages; the eight steps it summarizes are given here in full,
with their counts and with the rule sets, distance constants and per-branch
audits behind them.
\else
\textbf{Construction-step details (Section~\kcbsec{sec:kcb-construction}{3.2}).} The main text keeps the
eight-step ladder and its counts; the rule sets, distance constants and
per-branch audits behind the steps are reproduced here verbatim from the
pre-compression draft.
\fi

The pipeline turns ToolHop sub-questions into \dataset records in eight steps. Each step below states what it does and why, and reports how many questions survive it. Steps~2, 4 and 7 are performed by human annotators and are labelled \emph{manual}; Steps~1, 5 and 6 are executed by code and are labelled \emph{automatic}; Step~3 combines an automatic pass with manual adjudication. Every manual step is carried out by two annotators working independently.

\kcbstep{Step 1 (automatic)}{Flatten chains and deduplicate.}
This step establishes the unit of the benchmark. We take the sub-question, not the chain step, as the unit, because the same sub-question is reused across chains --- one formulation occurs in 17 different chains --- and step-level sampling therefore produces a pool with far less variety than its size suggests. We flatten all 3,912 sub-questions into a single list and collapse those whose normalized text is identical. \textbf{3,912 $\rightarrow$ 1,948 unique questions.}

\kcbstep{Step 2 (manual + automatic)}{Separate factual from computational questions.}
This step keeps only questions for which answering without a tool is meaningful, since a question whose answer is computed rather than recalled cannot create a conflict with parametric knowledge. A rule-based filter rejects a question when its answer is produced by an operation over an earlier answer rather than looked up: counting letters, reading ASCII codes, forming initials, extracting a name component, taking a difference between dates, comparing two values, or measuring a length. Because such a filter can over-reject, two annotators then manually reviewed all 54 questions whose rejection was not clear-cut and asked whether the rejection was justified; \textbf{19 were restored} --- a city's time zone, for instance, is a fact about a place rather than an arithmetic result. \textbf{1,948 $\rightarrow$ 605 factual questions} (1,343 computational questions set aside).

\kcbstep{Step 3 (automatic + manual adjudication)}{Control redundancy and leakage between questions.}
This step prevents two questions that share the same underlying fact from being split across a training and an evaluation fold, which would leak the answer. Exact duplicates are already removed in Step~1. A second automatic pass identifies near-identical formulations that differ only in a named argument (the time zone of Orleans, Charles and Rockland counties). A third, also automatic, pass identifies \emph{knowledge clusters}: linked questions about the same family or narrative in which the answer to one question becomes the main entity of another. These clusters cross ToolHop chains. 412 of the 605 factual candidates belong to a cluster of size greater than one, and two annotators manually adjudicated the 58 questions whose cluster membership was ambiguous. No question is removed at this step; its output is the grouping used for cluster bootstrap and for fold assignment. \textbf{605 $\rightarrow$ 605 questions}, now grouped.

\kcbstep{Step 4 (manual)}{Structural review of question wording.}
This step ensures that each retained question can be answered on its own and does not give away its answer. \textbf{We did not rewrite or regenerate any question: we preserved the original ToolHop wording, and self-containedness was a selection criterion rather than an editing operation.} ToolHop sub-questions are already phrased as standalone questions, so the review decides which of them to keep. Two annotators independently reviewed all 605 candidates and removed 4 questions that refer to an earlier step, 1 that contains its answer in the question text, 7 dynasty questions whose answer follows from an ordinal in a title (the father of the sixth Earl of Exeter is the fifth Earl), and 18 whose verified answer was disputed. The verified answers themselves were checked in three ways, two automatic and one manual: 397 were compared \emph{automatically} against Wikidata through the property identifier used by the question (329 exact matches, 11 partial, 3 fuzzy, 12 mismatches, 28 with no such claim, 14 with an unresolvable subject), 208 were checked \emph{automatically} by a web-browsing language model, and 74 anchor answers were checked \emph{manually} by the authors against primary sources. \textbf{605 $\rightarrow$ 575 questions} with fixed wording, executable calls, and verified answers.

\kcbstep{Step 5 (automatic + manual)}{Answer-aware distractor construction.}
This step produces, for each question, the incorrect value that a tool return will carry. A credible incorrect value has to match the question's domain and period as well as the answer type, which a draw from ToolHop's own pool of verified answers --- that is, some other question's gold answer of the same nominal type --- does not deliver. Each substitution is therefore constructed from the verified entity itself, through one of three branches; the manual review that established this is reported below.

\emph{(a) Wikidata branch (246 questions).} For person-valued answers we resolve the verified entity through the relation-specific Wikidata property used by the question rather than by name search --- the answer to a question about a person's father is resolved through property \texttt{P22} --- which prevents an unrelated namesake from being selected. Candidate distractors are drawn from other real entities matched on occupation (\texttt{P106}), gender (\texttt{P21}), year of birth (\texttt{P569}) and citizenship (\texttt{P27}), excluding the verified answer and every other true value of that property for the subject. A near distractor lies within 45 years of the verified person's birth year; a far distractor is separated by 150--600 years or by country. Candidates with more than 40 Wikipedia language editions are excluded, so that a model does not reject a substitution because the entity is exceptionally prominent rather than because the fact is wrong.

\emph{(b) Temporal branch (173 questions).} Dates, years and time zones are handled by a deterministic procedure. The substituted value differs from the verified answer but is rendered in the verified answer's own surface format, so that the evaluation does not conflate factual conflict with format recognition. Distance is set on the calendar: a near date is offset by 15--1{,}500 days and a near year by 1--15 years, while a far date is offset by 18{,}000--110{,}000 days and a far year by 80--400 years.

\emph{(c) Human-authored branch (149 questions).} The remaining questions are routed to human authors when the verified entity cannot be resolved unambiguously in Wikidata or the metadata needed for automatic matching are absent: works, historical figures without reliable country information, and people whose profession or period cannot be established automatically. Two authors independently proposed distractors for 152 questions. They proposed the same value for 46 and different values for 106. Because several incorrect values can be equally plausible, we do not treat disagreement as annotation error and do not report it as consensus: both proposals are retained, and the benchmark value is selected by a rule fixed before adjudication. This branch supplies context-sensitive substitutions that answer type alone cannot yield, such as replacing a character from \textit{Father Ted} with another character from the same series, or a pharaoh with another pharaoh. After structural filtering, 149 human-authored records remain.

\emph{Near and far.} All three branches share one rule: \textbf{the answer type is always preserved, a \emph{near} distractor must be indistinguishable from the verified answer in type and plausibility, and a \emph{far} distractor must be obviously different in period or country.} What realizes ``obviously different'' is branch-specific and is stated with each branch above --- birth-year distance or citizenship for entities, calendar distance for dates and years. The distinction is a property of the distractor pair, not of the question, and it is frozen in the released mapping rather than recomputed at run time.

Seven further records received targeted automatic repairs, giving \textbf{575 records}: 246 Wikidata, 173 temporal, 149 human-authored and 7 repaired, of which 426 are constructed automatically and 149 carry human-authored substitutions.

\kcbstep{Step 6 (automatic)}{Programmatic validity checks.}
This step catches malformed records before any human reads them. The checks are assertions in the build script and involve no human judgement: every distractor is non-empty; no distractor equals the verified answer; the near and far distractors of a question differ from each other; and no distractor coincides with a value recorded in Wikidata as an alternative true value of the property the question asks about. All 575 records pass. \textbf{575 $\rightarrow$ 575 records.}

\kcbstep{Step 7 (manual)}{Blinded semantic review.}
This step checks the properties that code cannot check --- whether a question makes sense on its own and whether its incorrect value is believable --- before any model is run. Two annotators reviewed all 575 question--answer--distractor records under five checks, blind to each other's labels: whether the question is self-contained, whether it leaks the verified answer, whether the verified answer is valid, whether the distractor has the required answer type, and whether the substitution is plausible. A blinded replication of 58 records was annotated a second time to measure reliability (Section 3.4 of the main paper). The review returns 503 \textsc{pass}, 39 \textsc{repair} and 33 \textsc{exclude} verdicts. A \textsc{repair} changes only the substituted incorrect value; question wording and verified answers stay frozen byte-for-byte, including 169 inherited capitalization typos, which we document rather than correct. Removing the \textsc{exclude} records leaves the shared analysis set. \textbf{575 $\rightarrow$ 542 questions.}

\kcbstep{Step 8 (automatic)}{Instantiation into tool conditions.}
This step turns each quality-controlled record into the episodes a model actually sees, so that what varies between conditions is the observed tool return and nothing else. Each retained question is instantiated with (i) a correct return containing the verified answer $g$; (ii) an incorrect return containing the near distractor $d_{\mathrm{near}}$; (iii) an incorrect return containing $d_{\mathrm{far}}$, where one is available; (iv) a structured tool-error payload; and (v) no tool return. The question, verified answer, executable call and available distractors are fixed across models. Conditions (i), (ii), (iv) and (v) exist for every question and are the four that make up the 6,504 episodes quoted in the main text ($542 \times 4 \times 3$); condition (iii) is available for only 463 of the 575 questions and is analysed separately. Each factual return condition is evaluated under three instruction-wording variants with the question text held fixed. \emph{NoopTool} is evaluated separately as an interface control.

Figure~\ref{fig:teaser} instantiates these conditions for one benchmark question. The correct and incorrect factual returns enter the correctness matrix in Figure 2 of the main paper; tool error and no tool are separate controls outside that matrix.

\textbf{Dataset composition (Section~\kcbsec{sec:kcb-evaluation}{3.3}).} Reproduced from the
pre-compression draft.

\paragraph{Chain provenance and position.} The 605 factual questions are attributed to 310 ToolHop chains: when the same question text occurs in several chains, it is assigned to the chain in which it first occurs. The remaining 685 chains are therefore \emph{not} free of factual questions --- we verified that all 995 chains contain at least one question from the pool --- their texts simply collapsed during deduplication. Multiple questions per chain are the norm rather than the exception: of the 310 contributing chains, 279 yield two \dataset candidates, 8 yield three and 23 yield one. We retain every eligible sub-question instead of selecting one representative per chain. Factual questions sit early in the reasoning chain: 298 are first-step questions, 288 second-step and 19 third-step. After the structural review of Step~4, the 575 retained questions come from 309 of those chains. Training folds are cut by whole chain (fold A: 287 questions from 154 chains; fold B: 288 from 155; zero intersection).

\paragraph{Topics.} ToolHop labels each chain with a domain, and a question inherits the label of the chain it is attributed to. Over the 575-question pool these labels reduce to 47 distinct topics, out of 62 in ToolHop as a whole. The distribution is long-tailed and biographical: history 118 questions (20.5\%), film 94 (16.3\%), genealogy 80 (13.9\%), time zones 44 (7.7\%), mathematics 26 (4.5\%), computing 19 (3.3\%), time conversion 17 (3.0\%), literature 15 (2.6\%), music 14 (2.4\%) and entertainment 12 (2.1\%). These ten topics carry 439 questions, 76.3\% of the pool; the other 37 topics contribute 136 questions, none with more than twelve. Two properties of this label matter when reading the results. It describes the chain rather than the individual question, which is why topics such as mathematics survive the factual filter of Step~2 at all --- a chain about a calculation can still contain a lookup step. And the concentration in biographical domains, with history, film and genealogy alone accounting for 292 questions (50.8\%), is what makes people the dominant answer type: the verified answers of the 575-question pool are 304 people, 111 dates, 50 places, 44 years, 23 organizations, 18 time zones, 17 other strings and 8 works. That mix is also why the Wikidata branch of Step~5 covers the largest share of the pool, and it bounds the domains over which our conclusions were measured.

\paragraph{Answer types.} ToolHop does not record an answer type per step: its \texttt{previous\_answer\_type} field is chain-level, covers five values (person 498, date 274, year 151, place 44, organization 28) and cannot be attached to a specific step when a chain branches. The category of each verified answer is therefore derived from its Wikidata \texttt{P31} (\emph{instance of}) property, which is what the distractor branches of Step~5 dispatch on. The automatically assembled type map this replaced, and the audit that retired it, are recorded below.

\paragraph{Coverage of the far condition.} A far distractor exists for 463 of the 575 questions (443 after the semantic exclusions of Step~7). It is missing for 105 human-authored records, for all 17 \texttt{other\_string} answers and for all 8 \texttt{work} answers. Near--far comparisons are therefore restricted to this subset, which is skewed towards automatically constructed items --- only 44 of the 149 authored records are covered --- and is not a random sample of the benchmark set.

\textbf{Superseded construction steps.} Two elements of the pipeline were
replaced during development. The main text describes only the final
construction; they are recorded here as provenance. \emph{Same-type sampling.}
The first procedure drew each incorrect value uniformly from ToolHop's own pool
of verified answers of the same nominal type --- that is, another question's
gold answer. A manual review of all 605 candidates rejected 167 of them:
approximately 106 name fragments, 40 substitutions of the wrong gender for the
relation asked about, and 9 strings containing part of the verified answer,
with the remainder type-compatible but implausible for the question's domain or
period --- a chemist replaced by a canal engineer. Surface defects can be
filtered automatically, contextual implausibility cannot, which is what
motivated resolving the verified entity itself and constraining candidates on
occupation, period and country. \emph{Answer-type map.} Those defects traced to
an automatically assembled type map: GLiNER (\texttt{gliner\_small-v2.1},
threshold 0.5) was run over the 807 unique normalized verified-answer strings,
its person, location, organization and creative-work labels mapped onto our
four categories by highest-scoring span, and ToolHop's own field preferred
where it applied, overriding 24 predictions. A later manual audit measured
category mismatches in 13\% of near distractors and 18\% of far distractors,
mostly genealogical relations routed into the place-and-organization branch.
The current pipeline derives the category from Wikidata \texttt{P31} instead.
\emph{Generator audits.} Two blinded sampled audits bracket the automatic
branch: annotator agreement $\kappa=0.53$ with 12\% of inspected distractors
rejected before the type-map fix, and $\kappa=0.67$ with the rejection rate
unchanged after it, 9 of the 11 remaining failures being pre-modern or
legendary figures without citizenship metadata.

\textbf{Scorer-v3 extension.} The regex extension for passive abstain
phrasings was validated against the same human labels with zero
over-fires on 120 QC answers and moves only the abstain channel:
retention, following, accuracy and incorrect-tool following stay within \textbar $\Delta$\textbar$\le$0.011 across all 15 arms.

\textbf{Scorer gate on the released benchmark (v1.3).} Before any
current metric was read, the deterministic scorer was validated
against blind human labels drawn from this benchmark's own responses,
under a pre-registered bar (outcome $\kappa$ $\ge$ 0.8; recall $\ge$ 0.9 on the
TOOL-followed and GOLD-kept classes). Three probes failed and drove
scorer iterations (quoted FINAL markers as mentions; prefixed
refusals; comma-headed aliases; question-echo, "is not known",
"did not provide", tool-re-run echoes; date-format equivalence) ---
each iteration regression-checked on all previously labelled rows.
The final enlarged probe (167 stratified rows, zero overlap with the
failed probes, population re-scored by v1.3 before sampling) passed:
$\kappa$ 0.806, TOOL recall 23/23, GOLD recall 31/33 = 0.939; the two
annotators agree at 161/167 ($\kappa$ 0.952), six disagreements adjudicated,
and the adjudicated conventions (agree-value answers are BOTH, not
GOLD; degenerate schema echoes, "does not have X in the provided
data", and refusal-FINALs anchor to ABSTAIN; year-only dates match
full dates bidirectionally) are frozen in the released packet README.
The disclosed blind zone: ABSTAIN recall 0.706 (long polite refusals
read as OTHER), which the judge layer targets (\S\kcbsec{sec:additional-analyses}{5.4}).

\textbf{CAR detector details (previous benchmark version).} Everything in this
paragraph --- the detector's precision, the two-stage estimator and its
per-model brackets --- was measured on the previous benchmark version, where
positive acknowledgments existed at all. It is reported as evidence about
the \emph{detectors}, not as an estimate of the acknowledgment rate: the
brackets sit above the direct measurement on the released benchmark, which is
0/120. The deployed detector is restricted to
tool\_wrong answers, where the LLM judge is more precise and its false
positives do not cluster on tool\_error; precision against human labels
is 0.38 on tool\_wrong (0.27 across all conditions). Two honesty notes:
the 11-positive validation set was reused across both judge
iterations, so judge validation numbers are in-sample, and prompt
iteration was stopped there to avoid fitting the detector to its own
test. Since judge over-firing plausibly correlates with answer
verbosity, no cross-model CAR claims are made from the judge screen
alone. In the two-stage estimator, the per-arm rate is recovered as
flag-rate $\times$ human-precision + (1$-$flag-rate) $\times$ human-miss-rate over
2,403 screened tool-wrong answers; the per-model brackets are gemma
{[}0.01, 0.05{]}, base Llama {[}0.02, 0.07{]}, Llama-Instruct {[}0.02, 0.08{]},
Mistral {[}0.05, 0.09{]}, Qwen {[}0.05, 0.16{]}. Rubric history: the first
rubric left open whether "the tool result does not match the question"
counts as acknowledgment ($\kappa$ 0.196); the second --- fixed before any
re-annotation --- requires the answer to reference the model's \emph{own}
knowledge, and flipped 44 of 77 previously agreed acknowledgments
($\kappa$ 0.351 at raw agreement 0.87, depressed by the now-rare positive
class).

\textbf{Desired-behaviour validation details.} Actions $\in$ \{follow tool, use
own knowledge, present both, abstain\} plus a binary must-flag axis,
under an enterprise-policy framing (the tool is an
authoritative-but-fallible internal database), blind to correctness
and cell. On the arbitration case neither annotator ever chose to follow the
tool (0/30 each). The must-flag axis agrees at 0.94 raw but $\kappa$ 0.237 --- a
rare-class $\kappa$ paradox (both annotators flag near-universally).

\textbf{Cell sizes (released benchmark).} The closed-book cell map is
fixed per (model, prompt formulation) by one forced elicitation and shared by
every arm of that model and formulation --- standard, presentation and prompting alike --- which is what makes the treatment
statistics exactly question-paired. Cross-formulation conflict
cases: arbitration 67--162, neither-correct 293--414 across the five models
(the correct-return case mirrors it at 293--414 --- both hold the
closed-book-wrong questions; tool\_error and no-conflict span all 542;
mem\_absent 2--125, excluded from conflict metrics); the judge's
memory-correctness relabel moves only the pretrained checkpoint (162$\rightarrow$136 --- echo-style
closed-book answers re-labelled). Its cases under Paraphrase~A and~B collapse
to n=18/10 (closed-book degeneration, \S\kcbsec{sec:default-arbitration}{5.1}) and are not reported.
The finetuning arms are cross-fitted, so each carries its own
out-of-fold support: the paired fine-tuning contrasts reported in the main paper run on 76--201
shared questions (184--445 formulation-pooled units) on the arbitration
case and 382--481 shared questions on the correct-return case. Because a cell is defined
by the arm's own parametric answer, treatment moves cell membership;
contrasts are computed on the intersection of supports and the drift
(questions entering and leaving) is reported next to every delta in
the released summary. The largest drift is on Mistral's and gemma's SFT arms in the
correct-return case, which is consistent with their large movements in
incorrect-tool following (\S\kcbsec{sec:tuning-results}{5.2}).

\section{Additional Stability Analyses}
\label{app:e}

\begin{figure}[htb!]
\centering
\begin{tikzpicture}[font=\scriptsize,
  ax/.style={draw=black!55, semithick},
  rng/.style={draw=black!18, line width=2.6pt, line cap=round},
  reference/.style={fill=black, draw=none},
  paraphraseA/.style={fill=white, draw=black, semithick},
  paraphraseB/.style={fill=black!55, draw=none, rectangle}]

\node[anchor=south west, font=\scriptsize\bfseries]
  at (33.5mm,23.6mm) {Correct-answer retention};

\node[anchor=east, align=right, text width=30mm]
  at (31mm,20.00mm) {Llama-3.1-8B-Instruct};
\node[anchor=east, align=right, text width=30mm]
  at (31mm,15.25mm) {gemma-2-9b-it};
\node[anchor=east, align=right, text width=30mm]
  at (31mm,10.50mm) {Qwen2.5-7B-Instruct};
\node[anchor=east, align=right, text width=30mm]
  at (31mm,5.75mm) {Mistral-7B-Instruct-v0.3};

\draw[rng] (41.93mm,20.00mm) -- (55.92mm,20.00mm);
\node[circle,reference,inner sep=1.0pt] at (41.93mm,20.00mm) {};
\node[circle,paraphraseA,inner sep=1.0pt] at (55.92mm,20.00mm) {};
\node[paraphraseB,inner sep=1.0pt] at (49.47mm,20.00mm) {};

\draw[rng] (37.91mm,15.25mm) -- (44.26mm,15.25mm);
\node[circle,reference,inner sep=1.0pt] at (37.91mm,15.25mm) {};
\node[circle,paraphraseA,inner sep=1.0pt] at (44.26mm,15.25mm) {};
\node[paraphraseB,inner sep=1.0pt] at (43.46mm,15.25mm) {};

\draw[rng] (36.39mm,10.50mm) -- (44.37mm,10.50mm);
\node[circle,reference,inner sep=1.0pt] at (44.37mm,10.50mm) {};
\node[circle,paraphraseA,inner sep=1.0pt] at (36.39mm,10.50mm) {};
\node[paraphraseB,inner sep=1.0pt] at (37.58mm,10.50mm) {};

\draw[rng] (39.28mm,5.75mm) -- (48.07mm,5.75mm);
\node[circle,reference,inner sep=1.0pt] at (42.83mm,5.75mm) {};
\node[circle,paraphraseA,inner sep=1.0pt] at (39.28mm,5.75mm) {};
\node[paraphraseB,inner sep=1.0pt] at (48.07mm,5.75mm) {};

\draw[ax] (33.5mm,-1.0mm) -- (59.0mm,-1.0mm);
\draw[ax] (33.50mm,-1.0mm) -- (33.50mm,-1.9mm);
\node[anchor=north] at (33.50mm,-1.9mm) {0};
\draw[ax] (42.61mm,-1.0mm) -- (42.61mm,-1.9mm);
\node[anchor=north] at (42.61mm,-1.9mm) {0.1};
\draw[ax] (51.71mm,-1.0mm) -- (51.71mm,-1.9mm);
\node[anchor=north] at (51.71mm,-1.9mm) {0.2};

\node[anchor=west] at (0mm,-6.6mm) {%
  \tikz\node[circle,fill=black,inner sep=1.0pt]{};\, Reference \quad
  \tikz\node[circle,fill=white,draw=black,semithick,inner sep=1.0pt]{};\, Paraphrase A \quad
  \tikz\node[fill=black!55,rectangle,inner sep=1.0pt]{};\, Paraphrase B};

\end{tikzpicture}
\caption{Correct-answer retention across three instruction-wording variants with fixed question text on untreated models. Markers show the Reference, Paraphrase A, and Paraphrase B estimates; gray segments show their minimum--maximum range and are not confidence intervals. The ranges overlap and model rankings change across formulations. Qwen2.5-7B-Instruct has the highest retention on the Reference formulation and the lowest on Paraphrase A, so we do not include cross-model retention rankings among the core claims.}
\label{fig:spread}
\end{figure}
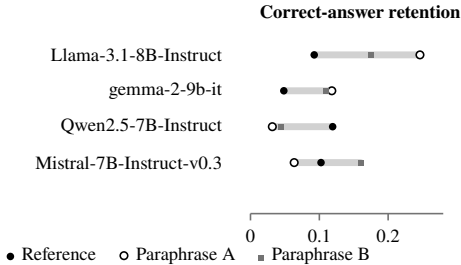

\ifnum\kcbtarget>1\relax\else%
\textbf{Distractor distance (Section~\kcbsec{sec:additional-analyses}{5.4}), full estimates.} The contrast is question-paired and read on the Reference wording, the only formulation for which both conditions exist; a far distractor exists for 463 of the 575 questions (443 after the semantic exclusions of Step~7), and the covered subset is skewed towards automatically constructed items, so it is not a random sample of the benchmark set. Replacing the near distractor with one separated by 150 to 600 years or by country for entities, and by a comparable margin for dates, raises correct-answer retention on Meta-Llama-3.1-8B from 20.6\% to 40.8\% ($+19.4$ points, 95\% CI $[9.7, 29.1]$, Holm-adjusted $p=0.0015$, $n=103$ questions), while all four instruction-tuned models are null ($|\Delta| \leq 4.2$ points, Holm $p=1.0$). When neither source is correct, every model repeats a distant incorrect value less often, from $-8.2$ points on the base model to $-19.7$ on gemma-2-9b-it (all $p \leq 0.003$). The deterministic and judge-normalized layers agree on the verdict for every model; on Meta-Llama-3.1-8B-Instruct they differ in sign while both are null, so we claim agreement of verdicts rather than of signs. On a separate axis, Figure~\ref{fig:spread} shows correct-answer retention under the three prompt formulations on untreated models: the ranges overlap and the model ranking reverses between them (\S\kcbsec{sec:formulation}{5.3}).

\fi%
\ifnum\kcbtarget>1\relax\textbf{Prompt-formulation spread.} Figure~\ref{fig:spread} shows
correct-answer retention under the three prompt formulations on untreated
models: the ranges overlap and the model ranking reverses between them
(\S\kcbsec{sec:formulation}{5.3}).\fi

\textbf{Unpaired base-versus-instruct retention family (Reference formulation).} The
per-cell two-proportion family (planned, Holm-corrected; cell sizes as
as reported below) agrees with the question-paired base-vs-instruct design on three
of four pairs: base vs gemma Holm-p 0.0007, vs Llama-Instruct 0.017 and
vs Mistral 0.041, but not vs Qwen (0.129, on the smallest arbitration case,
n=67). The paired family, which is the pre-registered one, separates all
four.

\textbf{Prompting, all four models and all three strategies (\S\kcbsec{sec:additional-analyses}{5.4}).}
Table~\ref{tab:prompting} gives the levels behind the prompting result. One
sentence changes per arm and nothing else; because the arms of a block share
that model's own closed-book answers, differences within a block are exactly
question-paired.

\begin{table}[t]
\centering
\setlength{\tabcolsep}{3pt}
\renewcommand{\arraystretch}{1.05}
\small
\resizebox{\linewidth}{!}{%
\begin{tabular}{@{}llcccc@{}}
\toprule
\textbf{Model} & \textbf{Prompt}
& \textbf{Ret.}$\uparrow$
& \textbf{Tool}$\uparrow$
& \textbf{Wrong}$\downarrow$
& \textbf{Err.}$\uparrow$ \\
\midrule

\multirow{4}{*}{Llama-3.1-8B-Instruct}
& --               &  9.3 & 87.8 & 71.4 &  89.3 \\
& Warning          & 21.0 & 75.1 & 65.3 &  16.4 \\
& Source-priority  & 50.6 & 79.4 & 60.9 &  29.7 \\
& Abstain-and-flag & 12.3 & 23.5 & 18.8 &  89.7 \\
\midrule

\multirow{4}{*}{gemma-2-9b-it}
& --               &  4.8 & 80.2 & 74.1 & 100.0 \\
& Warning          & 17.7 & 80.2 & 65.5 &  89.7 \\
& Source-priority  & 25.0 & 82.6 & 75.1 &  81.9 \\
& Abstain-and-flag & 21.0 & 58.4 & 39.2 &  97.0 \\
\midrule

\multirow{4}{*}{Qwen2.5-7B-Instruct}
& --               & 11.9 & 82.6 & 75.4 &  88.8 \\
& Warning          & 40.3 & 70.0 & 56.5 &  11.4 \\
& Source-priority  & 37.3 & 75.1 & 62.6 &  22.5 \\
& Abstain-and-flag &  1.5 & 25.1 & 18.8 & 100.0 \\
\midrule

\multirow{4}{*}{Mistral-7B-Instruct-v0.3}
& --               & 10.2 & 88.2 & 82.9 &  75.6 \\
& Warning          &  8.7 & 88.2 & 78.2 &  23.1 \\
& Source-priority  & 18.9 & 85.8 & 74.7 &  58.5 \\
& Abstain-and-flag & 43.3 & 35.3 & 26.8 &  52.8 \\
\bottomrule
\end{tabular}
}
\caption{Prompting interventions on the four instruction-tuned models,
Reference formulation. A dash denotes the same checkpoint under the standard
evaluation prompt. Ret., Tool, Wrong and Err. are as in Table~1 of the main
paper: correct-answer retention, correct-tool following, incorrect-tool
following and tool-error abstention. These are single-formulation levels and
therefore differ from the formulation-pooled rows of Table~1; the paired
changes with intervals and Holm tests, pooled over the three formulations, are
in the released summary layer. Retention, correct-tool following and
incorrect-tool following are judge-normalized, tool-error abstention is a
deterministic-scorer value.}
\label{tab:prompting}
\end{table}

\emph{Reading.} Every model gains retention from at least two strategies
(eleven of twelve pooled comparisons are Holm-significant, \S\kcbsec{sec:additional-analyses}{5.4}) while
paying a model-specific cost. Llama-Instruct and Qwen give up correct-tool
following under the warning and source-priority prompts; gemma and Mistral
preserve it there but still lose tool-error abstention. Abstain-and-flag
collapses correct-tool following on every model and is Qwen's one null, with
retention at 1.5. gemma is the only model whose tool-error abstention survives
the warning prompt nearly intact, 100.0 to 89.7.

\ifnum\kcbtarget>1\relax\else%
\textbf{The one-sentence protocol ablation (\S\kcbsec{sec:formulation}{5.3}).} The standard
with-tool prompt carries a protocol sentence --- "the tool cannot be
called again --- answer now" --- that has no epistemic content about
sources. We remove it and rerun the anchor model on the full
released benchmark (identical episodes, temperature 0, question-paired,
judge-normalized). On the arbitration case the sentence does
\emph{nothing}: retention changes by $-0.019$, 95\% CI $[-0.068, +0.031]$,
$p=0.65$, $n=162$, and the deterministic layer agrees ($-0.025$,
$p=0.43$). What it does move is obedience where the tool is right:
following a correct tool return falls by 0.093 $[0.048, 0.138]$
($p=0.0002$, $n=378$) and no-conflict accuracy by 0.056 ($p=0.011$);
tool-error abstention is borderline and we read it as null ($-0.033$,
$p=0.069$). The sentence buys compliance, not deference.

On the previous benchmark version's validation slice the same ablation was measured at
roughly 25 points on the arbitration case (retention 0.75$\rightarrow$0.47,
incorrect-tool following 0.16$\rightarrow$0.41, tool-error abstention
0.23$\rightarrow$0.52), and was described in earlier drafts as the largest
single-intervention movement we had measured there. It does not survive
the released benchmark. We record the discrepancy rather than dropping
it: it is one of the construction-scope effects this paper reports, and
the earlier estimate was taken on a slice whose arbitration case was both
smaller and differently composed.

\fi%
\textbf{Seed robustness of the LoRA effect (\S\kcbsec{sec:tuning-results}{5.2}).} The primary adapters
use seed 20260705; two further seeds (42 and 20260719) were trained and
evaluated under the identical cross-fitted protocol on the two anchors
that meet the criterion. Judge-normalized, formulation-pooled change in
retention on the arbitration case:
Llama-Instruct +0.058 / +0.077 / +0.086, gemma +0.093 / +0.111 /
+0.140; all four additional arms are Holm-significant at p 0.0004
within family T3, all three seeds agree in sign per model, and the
per-model CIs share a common interval ({[}+0.056, +0.088{]} on the anchor,
{[}+0.096, +0.131{]} on gemma). The gain is therefore not a single-seed
artifact. The two models that \emph{fail} the criterion were not seed-replicated (\S\kcbsec{sec:limitations}{6}).

\textbf{Per-formulation verdicts of the asymmetric criterion (\S\kcbsec{sec:tuning-results}{5.2}).} The
criterion (retention rises at Holm $p<0.05$ with no significant fall in
correct-tool following) was pre-registered on the pooled slice and, separately, on
each prompt formulation. Pooled it is met by Llama-Instruct and gemma under
both losses. Per formulation, in Table~\ref{tab:pvcrit}:

\begin{table}[t]
\centering
\setlength{\tabcolsep}{3pt}
{\small
\begin{tabularx}{\columnwidth}{Xcccc}
\toprule
Arm & Reference & Par.~A & Par.~B & Pooled \\
\midrule
Llama-3.1-8B-I $\cdot$ SFT & met & fail$^{t}$ & met & met \\
Llama-3.1-8B-I $\cdot$ DPO & fail$^{k}$ & fail$^{k}$ & met & met \\
gemma-2-9b-it $\cdot$ SFT & met & met & met & met \\
gemma-2-9b-it $\cdot$ DPO & fail$^{k}$ & met & met & met \\
Qwen2.5-7B-I $\cdot$ SFT & fail & fail & fail & fail \\
Qwen2.5-7B-I $\cdot$ DPO & fail & fail & fail & fail \\
Mistral-7B-I $\cdot$ SFT & fail & fail & fail & fail \\
Mistral-7B-I $\cdot$ DPO & fail & \textbf{met} & fail & fail \\
\bottomrule
\end{tabularx}}
\caption{The asymmetric criterion evaluated per prompt formulation and pooled
(judge-normalized layer; Holm correction inside each slice's own family).
$^{k}$ failed because the retention gain does not reach Holm $p<0.05$
(0.28, 0.051 and 0.44 respectively); $^{t}$ failed on a significant
loss in correct-tool following ($-$0.045, $p$ 0.028). The deterministic layer agrees
except on Llama-Instruct's DPO arm under Paraphrase~A, which it reads as
met. Only
gemma's SFT arm meets the criterion under every formulation; Mistral's DPO
arm, a failure when pooled, meets it under Paraphrase~A.}
\label{tab:pvcrit}
\end{table}

This is the same instability the stability rule of \S\kcbsec{sec:formulation}{5.3} detects in
the untreated orderings, which is why \S\kcbsec{sec:tuning-results}{5.2} reports the 2/4 result as a
pooled finding whose passing set is formulation-scoped. The prompting
family behaves the same way: of the eleven pooled-significant retention
gains in the four-model $\times$ three-strategy matrix, only Llama-Instruct's
warn arm, gemma's source-priority arm and Mistral's abstain-and-flag
arm hold on all three formulations separately, and Mistral's warning arm is
negative under the Reference formulation ($-$0.016) against $+$0.065 pooled.

\ifnum\kcbtarget>1\relax\else%
\textbf{Switch compositions (previous benchmark version).} An oracle
prompt-switch inherits the knob it turns, and a realizable,
observables-only switch captures most of that ceiling, bounded by the
conflict cells in which no elicitable memory exists to disagree with
the tool. On gemma the oracle switch buys +0.380 accuracy on the
arbitration case over the standard-prompt arm (decided-only Y +0.295 {[}+0.198, +0.402{]}) and the
realizable switch +0.350 (Y +0.254 {[}+0.155, +0.352{]}), leaving a gap of
only +0.041 {[}$-$0.003, +0.092{]} between what an oracle could do and what
observable signals actually achieve; on the anchor the whole family is
indistinguishable from zero (oracle +0.091 accuracy, Y +0.050 {[}$-$0.053,
+0.158{]}; realizable +0.070, Y +0.041 {[}$-$0.047, +0.130{]}), because the
knob it composes is itself null on that model and formulation. The
switch's safety property is exact on both anchors --- zero false fires on
the agree cell --- and its structural ceiling is the 41--42\% of conflict
cells the observable gate cannot see at all, where no elicitable memory
exists to disagree with the tool. Any observables-based policy inherits
that ceiling.

\fi%
\ifnum\kcbtarget>1\relax\else%
\textbf{Formulation-spread composition block (previous benchmark version).}
Re-running the prompting arms and recomputing every composition under
the other two protocol formulations separates what is structural from
what the knob contributes. This block is the one composition not
re-run on the curated version (its control arms exist only under the
Reference formulation there); the Reference-formulation pattern it
brackets reproduced on the curated version. The switch's
\emph{safety} properties replicate exactly: zero false fires on the agree
cell under all three prompt formulations of both anchors, and the observable
gate's blind spot sits at 40--42\% of conflict cells everywhere. The
\emph{gains} inherit the prompt knob's formulation sensitivity: on gemma the
oracle buys +0.11--+0.31 decided-only accuracy across formulations
(always CI-positive) and the realizable switch +0.10--+0.28 (gap to
the ceiling 0.01--0.04); on the anchor the knob is indistinguishable
from zero under one formulation (oracle $\Delta$ $-$0.02 {[}$-$0.09, +0.06{]}) and
strong under another (+0.26 {[}+0.17, +0.36{]}), where the realizable
switch captures only about half the ceiling (gap +0.12 {[}+0.05,
+0.18{]}). The 40--42\% of conflict cells invisible to the observable
gate (no elicitable memory to disagree with) are a structural ceiling
for any observables-based policy.

\fi%
\textbf{OOD label provenance (Appendix~\kcbsec{app:f}{F}.1).} The full external-set label layer (981
counterfactual + 500 unanswerable rows) was produced by two LLM labeling
passes --- Claude Opus 4.8 Extra High (A) and ChatGPT/GPT-5.6-sol (B) --- with
prompts, decoding settings and request counts released; it is \emph{not} a human
double annotation, and we compute no human--human agreement for it. Human
evidence for this branch is confined to the separate 300-row human pilot
audit (which defines the verified 279/185 deployment pools) and to the scorer
gate (which validates the matcher, not item semantics).

\emph{The two passes are not independent evidence.} Both were run on 2026-07-19,
after the joint human adjudication of 2026-07-18, from a task file whose
two "previous annotator" columns carried the human labels in plain view. The
consequence is measurable. The human
pilot drew 300 of 1,000 rows at random (subsample share 0.26--0.34 across
deciles of the join order, expected 0.294), so the 294 rows humans inspected
and the 687 they did not are exchangeable and must carry the same defect
rate. Adjudication confirmed 15 gold-non-uniqueness defects inside the
inspected rows --- at least 5.1\%, hence about 35 expected among the 687. Each pass
flags exactly those 15 rows and 0 (A) / 1 (B) outside them; binomial
p = 2.4e-16 and 9.0e-15. Neither pass flags any of the 11 rows adjudication
cleared, nor any of the 83 rows one annotator flagged that adjudication never
reviewed. The passes replay the adjudicated verdict where it exists and
answer "clean" elsewhere. Their 966/981 mutual agreement is therefore
trivially high and is \textbf{not} a quality signal; the 687-row extension is
unverified by anything. Nothing here disturbs the Appendix~\kcbsec{app:f}{F}.1 estimates, which are
computed on the human-verified 279 and 185 pools.

\emph{A blind re-run was attempted and did not qualify.} We rebuilt the task
blind --- items only, hint columns and adjudication access removed, IDs
tokenized, rows shuffled --- over 494 rows (all 294 humans had inspected
plus 200 drawn from the 687 they had not), under a threshold fixed
before any label: recall $\ge$ 0.60 against the 15 jointly adjudicated
defects with $\le$ 0.30 false flags. The blinding sanity check passed ---
flag rates 0.218 inside the human-inspected part against 0.190 outside
(Fisher p = 0.50), against 15 versus 0--1 for the label-visible passes,
which independently confirms the diagnosis above --- but the utility
threshold failed: recall 0.400 {[}0.198, 0.643{]}, false flags 1/11. The
extended pool therefore cannot be rescued by an LLM screen, and every
reading stays on the human-verified pools. Two by-products are worth
recording: the blind pass flags 19.0\% {[}14.2, 25.0{]} of the 200
uninspected rows against a 5.1\% adjudicated defect rate in the
inspected part --- a flag is not a confirmed defect, but this is the only
quantitative signal we have about the extended pool --- and on the
contested \texttt{context\_\allowbreak{}supports} criterion the blind pass lands \emph{between}
our two annotators (19.6\% against 0\% and 28\%), so the stricter reading
is a systematic property of FaithEval's soft support rather than one
annotator's idiosyncrasy, which is why the passage frame is reported as
a sensitivity reading throughout.

\textbf{OOD pool construction (Appendix~\kcbsec{app:f}{F}.1).} The serialization filter removed
19/1000 label-desynced rows (answerKey not among the choice labels);
the human audit of a 300-row subset surfaced 21 defects (adjudicated
jointly), in two families: a programmatically filterable
serialization artifact and gold-non-uniqueness, which motivates
per-item uniqueness checks for any wholesale use of the source set.
Closed-book elicitation over the 981 filtered questions fixes each
anchor's memory-right set (819/981 anchor, 890/981 gemma, on the
revised matcher of Appendix~\kcbsec{app:f}{F}.1); both frames are read on the common
185-pool support (n=158/169 memory-right episodes per model), the
value frame additionally on the 279-pool as secondary (n=240/256). The abstention axis uses FaithEval's unanswerable split
(500 questions); closed-book retention is measured on the full
elicitation set.

\ifnum\kcbtarget>1\relax\section{Out-of-Distribution Transfer}\else\section{Out-of-Distribution Transfer and the 70B Scale Point}\fi
\label{app:f}

This appendix carries the two result families moved out of the main
text for length. Both are summarized in \S\kcbsec{sec:additional-analyses}{5.4}.

\subsection{Out-of-distribution transfer}

\textbf{An out-of-distribution transfer test finds the training gains
are not a construction signature --- they transfer, but partially,
attenuated, and frame-dependently, and which frame carries the
transfer depends on the loss.} The strongest objection to any
in-benchmark control result is that the model learned to distrust the
construction's own signature rather than arbitration-by-correctness.
We test this on an external conflict set built from FaithEval's
counterfactual split~\cite{ming2025faitheval} joined back to ARC-Challenge for
world-true gold: multiple-choice science questions whose
counterfactual answer comes from authored evidence --- a different task
format, domain and conflict provenance from everything in the training
pairs. The deployment pools are 279 value-frame and 185 passage-frame
episodes: the same counterfactual answer presented either as a bare
tool return value or as the full passage inside the tool return (pool
construction and supports in Appendix~\kcbsec{app:e}{E}). Because the MC matcher is a
new scorer domain, arm metrics were embargoed behind blind
two-annotator scorer gates, and all four pre-registered thresholds
passed ($\kappa$ 0.975 between human consensus and scorer, $\kappa$ 0.987 between
annotators, stratified abstention recall 1.00 at precision 1.00). The
enriched stratum of the second gate also forced a matcher revision,
disclosed below.

\emph{Label provenance.} The 981 counterfactual plus 500 unanswerable rows
were labeled in two LLM passes, not by independent human double
annotation, and a provenance audit shows the passes were run
with the human pilot's labels visible: they reproduce that pilot where
humans looked and answer "clean" almost everywhere else, flagging 0
and 1 of the 687 rows no human inspected against about 35 expected at the
audited 5.1\% defect rate. The 687-row extension is therefore
\textbf{unverified}, every reading below stands on the human-verified
279/185 pools, and the human gates validate the matcher and scorer,
not the semantic validity of every item.

The pre-registered readings (paired episode bootstrap, B=2,000), on
adapters trained leakage-free over the full 575-question pool, are given in Table~\ref{tab:ood}:

\begin{table*}[t]
\centering
\setlength{\tabcolsep}{4pt}
{\small
\begin{tabularx}{\textwidth}{Xcccc}
\toprule
Arm & Value frame & Passage frame & Abstention & Retention \\
\midrule
Llama-3.1-8B-Instruct $\cdot$ SFT & +0.051 & \textbf{+0.044} & $-$0.068 & +0.025 \\
Llama-3.1-8B-Instruct $\cdot$ DPO & \textbf{+0.070} & 0.000 & $-$0.020 & +0.017 \\
gemma-2-9b-it $\cdot$ DPO & \textbf{+0.112} & \textbf{+0.047} & $-$0.076 & +0.004 \\
\bottomrule
\end{tabularx}}
\caption{Out-of-distribution transfer (FaithEval-counterfactual
$\times$ ARC), adapters trained leakage-free over the full 575-question pool,
paired episode bootstrap $B=2{,}000$. Every cell is a change against the
corresponding original checkpoint, with its 95\% interval; bold marks an
interval excluding zero. The two conflict frames present the same
counterfactual value either as a bare tool return or inside a passage, and both
are read on the common 185-item support so that they are comparable. The 95\%
intervals are given in the text below.
Abstention is measured on the unanswerable split. Retention is the change in
closed-book accuracy, and all three arms fall inside the preregistered
$\pm0.03$ band.}
\label{tab:ood}
\end{table*}

The 95\% intervals, in the order of the table: value frame $[-0.019, +0.127]$, $[+0.006, +0.133]$ and $[+0.065, +0.166]$; passage frame $[+0.006, +0.082]$, $[-0.038, +0.038]$ and $[+0.012, +0.083]$; abstention $[-0.096, -0.038]$, $[-0.038, -0.002]$ and $[-0.102, -0.052]$.
Both conflict frames are read on the common 185-item support (n =
158 memory-right items on the anchor, 169 on gemma) so that the two
frames are directly comparable; on the wider 279-item value pool the
same arms give +0.038, +0.046 and +0.086. Retention is the closed-book
accuracy change on the 981-item elicitation set against a
pre-registered $\pm$0.03 band.

Transfer is real but \textbf{frame-scoped, and the frame depends on the
loss}: the anchor's DPO arm transfers in the value frame only, its
SFT arm in the passage frame only --- opposite frames --- while gemma's DPO
transfers in both, so the presentation frame is a first-order axis
even for a trained effect, echoing \S\kcbsec{sec:additional-analyses}{5.4} from the treated side.
\textbf{The abstention leak is confirmed on all three arms}, not on SFT
alone as the previous benchmark version suggested, making it the most consistent
side effect we measure anywhere. It is also where the in-distribution
reading must be reconciled: the anchor's DPO arm shows \emph{no} detectable
tool-error leak inside \dataset (+0.014, p = 0.062, \S\kcbsec{sec:tuning-results}{5.2}) and still loses
abstention here ($-$0.020). These are different measurements --- abstaining
on a broken tool call versus on an unanswerable question in a new
domain --- and both stand; the abstention axis is the fragile one on
seven of our eight in-domain arms and on all three transferred arms.
Per-frame nulls remain power-bounded (minimal detectable effect
0.11 to 0.16), not demonstrated absences, and every positive reading on
the anchor's DPO arm is a narrow interval we report as directional.
Transfer also runs in the \emph{reverse} direction. Adapters trained by
correctness-labelled LoRA-DPO on FaithEval/ARC rows only --- with zero
ToolHop content of any kind --- and evaluated on the untouched \dataset benchmark set
raise kept memory on gemma by $\Delta$ +0.046 {[}+0.025, +0.072{]} (Holm p 0.0008,
n = 158 shared questions) with the right-tool leg intact (+0.010), and
leave the anchor null (+0.012 {[}$-$0.009, +0.030{]}, n = 197). Because the
training data shares no construction, no question pool and no distractor
mapping with \dataset, this is the cleanest available evidence that the
trained behaviour is not a signature of our construction --- but it holds
on one of two models, so it supports a model-dependent replication
rather than a dataset-independent property of the recipe.

\textbf{Disclosure: the MC matcher was revised after the arms were read.}
The enriched gate stratum exposed a failure mode the pre-registered
thresholds did not cover: where the matcher matched an answer to no
option, both annotators nonetheless read an explicit option in 67--73\%
of cases, and the unmatched fraction differed across arms (7.65\% on
one untreated arm against 0.71\% on its SFT arm), so the miss biased
\emph{deltas} and not only levels. Following the pre-registration's remedy
--- corrections by offline rescoring only, never by regenerating
responses --- we repaired the matcher and rebuilt both the new and the
legacy summaries on it; the tables above are the rebuilt values. What
mitigates this: the direction of the correction was predicted before
rescoring from the differential unmatched rate; an independent earlier
gate on different arms and episodes, annotated before this rule
existed, is unchanged at 40/40; gemma, whose unmatched rate was
already near zero, moved by exactly 0.0000; and the rule's thresholds
are flat across the coverage $\times$ margin grid we swept. What it does not
undo: the repair \textbf{flipped one retention gate from fail to pass} (the
anchor's SFT arm, +0.045 $\rightarrow$ +0.025 against a $\pm$0.03 band), which is why
that arm is readable at all above, and the residual bias points the
same way. Residual matcher error on the gate is 4/55. Full tables and
the replication script are in the released provenance for this run.

\textbf{OOD transfer scope.} The out-of-distribution test is one
external set, multiple-choice and science-domain, so format transfer is
tested but free-form transfer is not; its verified conflict pools are
small, per-frame nulls are power-bounded (minimal detectable effect 0.11 to
0.16) rather than
demonstrated absences, and several positive readings are narrow
intervals we report as directional. All OOD claims are restricted to
the human-verified 279/185 pools because the 687-row extension is
unverified (Appendix~\kcbsec{app:e}{E}). Transferred magnitudes are direction-, not
size-preserving.

\ifnum\kcbtarget>1\relax\else%
\subsection{The 70B scale point}

\emph{A 70B scale probe (one model, one quantization, no
seeds).} Llama-3.3-70B-Instruct, run as a public AWQ-int4
requantization with its own fresh elicitation on the same benchmark set, buys a
real but partial improvement, unevenly across axes. It defends correct
memory better than any 8--9B instruction-tuned model (retention on the
arbitration case 0.21 formulation-pooled against their 0.07--0.17; paired
against Llama-Instruct on
common support, $\Delta$ +0.133 {[}+0.076, +0.196{]}) and follows the right tool
most (correct-tool following 0.96 pooled, $\Delta$ +0.029). That contrast
carries a large support drift --- 401 of the 70B's arbitration questions have
no counterpart
in the anchor's cell and 48 the reverse, so the pairing runs on the 185
questions where both models hold the correct answer. The other two axes
move the wrong way: with no correct memory to defend it is \emph{more}
deferent (incorrect-tool following 0.85 pooled, $\Delta$ +0.068),
and it is markedly worse at refusing a broken tool (tool-error
abstention 0.49 under the Reference formulation against 0.76--1.00 for the
8--9B instruction-tuned models,
$\Delta$ $-$0.410). The reading is a trade, not a lift: at 0.21 pooled retention this
model still abandons its own correct answer in roughly four of five
conflict cells, beating the anchor's 0.17 pooled but nowhere near
solving the problem. Scale is confounded here with model generation and
quantization, and it is a single point, so we read it descriptively.

\fi%
\section{Tool-Choice Protocols in an Executed Agent Loop}
\label{app:g}

This appendix carries the dev-stage probe referenced in \S\kcbsec{sec:additional-analyses}{5.4}. It supports
exactly one claim there --- that reliability is a model $\times$ protocol
interaction, and that constraining the decoder to the schema repairs the
Llama failure --- and nothing beyond it. The remaining rows are reported
\ifnum\kcbtarget>1\relax as a bounded result on an axis the benchmark
itself does not vary.\else for the same reason as the SAE-steering probe, as
a bounded result on an axis the benchmark itself does not vary.\fi

\textbf{Setup.} We replay the memory-right slice of this benchmark set --- the
162 question pairs (324 episodes) on which the evaluated model's forced
closed-book answer is correct --- through a live agent runtime in which
the model emits a tool call over its own native function-calling
template, the call is executed, and the (overwritten) return is fed back.
The runtime is vLLM 0.8.5.post1 with native tool calling enabled and the
per-model tool-call parser, so the protocol effects below are measured as
a standard serving stack realizes them.
Only the tool-choice protocol varies: \emph{a1}, a named schema-guided
call; \emph{a2}, free \texttt{auto} choice; \emph{a4}, a two-stage
decide-then-call prompt. Decoding is greedy. Intervals are percentile
bootstrap, $B=2{,}000$, over the per-model memory-right slice. Table~\ref{tab:agentic} reports the arms.

\begin{table}[t]
\centering
\setlength{\tabcolsep}{4pt}
{\small
\begin{tabularx}{\columnwidth}{Xcccc}
\toprule
Model & Guided & Auto & Two-stage & $n$ \\
\midrule
Qwen2.5-7B-Instruct       & 1.000 & 1.000 & 0.984 & 61 \\
Qwen2.5-3B-Instruct       & 1.000 & 0.979 & 0.979 & 48 \\
Qwen2.5-1.5B-Instruct     & 1.000 & 0.970 & \textbf{0.576} & 33 \\
Llama-3.1-8B-Instruct     & 0.992 & \textbf{0.328} & 0.976 & 125 \\
Llama-3.2-3B-Instruct     & 1.000 & \textbf{0.291} & \textbf{0.367} & 79 \\
Mistral-7B-Instruct-v0.3  & \textbf{0.104}$^{\dagger}$ & 0.885 & \textbf{0.250}$^{\dagger}$ & 96 \\
Hermes-3-Llama-3.1-8B     & 1.000 & 0.964 & 1.000 & 110 \\
\bottomrule
\end{tabularx}}
\caption{Answer agreement with the executed tool return on the
memory-right slice, by model and tool-choice protocol. The three protocols are
a named schema-guided call, free \texttt{auto} choice, and a two-stage
decide-then-call prompt. Bold marks a failing arm. $^{\dagger}$Mistral's two
schema-guided arms are confounded with a serving defect described below and
must not be read as a model property. Hermes-3 is a third-party tool-finetuned
checkpoint of the same Llama-3.1-8B base, included as an uncontrolled
comparison.}
\label{tab:agentic}
\end{table}

\textbf{Reading.} Which protocol fails depends on the model \emph{and} its
scale --- Llama under \texttt{auto}, Mistral under the guided call,
Qwen2.5-1.5B under the two-stage prompt --- and the failures are
mechanically distinct rather than one bug. Four mechanisms appear in the
traces: (T1) argument type coercion, in the Llama family under
\texttt{auto}, 458 type errors dominated by booleans serialized as
strings; (T2) unrenderable arguments, in Mistral's guided arms, where
302 of 302 rejected calls fail to parse as JSON at all; (T3) stage-one
decision-parse failure, in the two-stage protocol; and (T4) silent
abstention from calling the tool. Two repairs bracket T1: preference
fine-tuning on our own pairs does not remove it (444 type errors
remaining, $0.328 \rightarrow 0.360$), whereas constraining the decoder to
the schema does ($0.328 \rightarrow 0.992$) and a third-party tool-finetuned
checkpoint shows 14 type errors against 458, with no stringified values
at all. The controlled repair we own is the schema-constrained decoder;
the tool-finetuned checkpoint is an observation, not an ablation.

\paragraph{Metric caveat.} The score above is answer agreement: it matches
the final answer against the tool's value. On the memory-right slice the
tool's value and the model's own memory coincide, so an episode that
never calls the tool can still score. Against the tool-execution rate
recovered from the traces, 16 of 21 arms agree to within 1.5 percentage
points --- the collapses above are unaffected, and on every Llama arm the
two quantities are equal to three decimals --- but two do not: Mistral
under \texttt{auto} scores 0.885 while executing the tool in 0.385 of
episodes, and Qwen2.5-1.5B under the two-stage prompt scores 0.576 while
executing the tool in \emph{none}. We therefore report both quantities,
and read the small-model two-stage failure as silent abstention rather
than as malformed calls. In particular, Mistral does not "recover" under
\texttt{auto}: it mostly does not call the tool.

\paragraph{Reproducibility and sampling.} Decoding is greedy, so a
second run under a different seed emits byte-identical tool calls in 324
of 324 episodes. This establishes that the pipeline reproduces end to end;
on its own it is not evidence about sampling. We therefore re-ran the memory-right slice with the closed-book elicitation held greedy---which fixes the evaluation subset---and every tool-choice arm sampled at $T{=}0.7$, on three seeds. The collapse persists: Llama-3.1-8B follows the named schema-guided call on $1.000$ of episodes and drops to $0.268$, $0.268$ and $0.323$ under \texttt{auto} ($\Delta$ $-0.732$, $-0.732$, $-0.677$; mean $-0.714$, sd $0.032$; every 95\% interval excludes zero), against $-0.664$ $[-0.744, -0.576]$ under greedy decoding, while Qwen2.5-7B stays flat ($+0.000$, $+0.016$, $+0.000$). Sampling did vary the runs---per-episode outcomes differ on 30 to 40 of 324 episodes across seed pairs---so the aggregate is stable while individual episodes are not. Two caveats: the sampled runs score 127 memory-right episodes rather than 125, because greedy elicitation is not bit-stable across runs on this server (two earlier greedy runs already differ from each other by three episodes, while the three sampled seeds share an identical subset); and this is a single temperature, not a curve.

\paragraph{Serving confound on the guided arms.} Under the vLLM version
used here, Mistral's native tokenizer path hard-errors on the guided
render and the fallback path soft-fails, so its guided-arm figure is a
property of the serving stack and not of the model. The same class of
defect produced one unusable Qwen2.5-7B cell under a prompt formulation, where the
constrained decoder looped on JSON whitespace and 238 of 260 calls never
parsed, while the same model on the same episodes under \texttt{auto}
executed 260 of 260 cleanly. Guided-arm numbers are reported with this
caveat attached, and the free-choice arm is used as the clean comparison
point.

We tried to isolate the defect by re-running the affected arms under a
different constrained-decoding backend, and report the attempt because it
failed informatively. On this vLLM version the backend selector is inert:
switching it reproduced the unusable cell to the episode (238 of 260
rejected either way, and 42 of 260 on the two-stage arm), and the only
other selectable value is the one the default already resolves to, so the
implementation cannot be varied here at all. The failure is therefore
established as deterministic rather than flaky, but its attribution to the
decoder rests on the cross-arm and cross-formulation evidence above, not on
a backend swap. \textbf{This does not reach the \S\kcbsec{sec:additional-analyses}{5.4} claim}: the arm that
collapses there is free \texttt{auto} choice, which sets no schema and
invokes no constrained decoder, so that collapse cannot be an artifact of
one. The same re-run reproduced it independently, at
$0.984 \rightarrow 0.312$, $\Delta$ $-0.672$ $[-0.752, -0.584]$.

\ifnum\kcbtarget>1\relax\else%
\section{Whether the Model's Own Answer Is Still Available}
\label{app:h}

A natural question about the arbitration case is whether the model's own
correct answer is still available to it at the moment it adopts the wrong
tool value, or whether the tool return displaces it. This appendix reports
a dev-stage probe of that question. It supports no claim in the main
paper: it covers two model families of four, one construction, and one
reading position, and the quantity it measures is a property of the
representation, not of behaviour.

\textbf{Measure.} We read the residual stream through a Jacobian lens at
the final input position, and record the rank of the row corresponding to
the model's own elicited closed-book answer in the resulting vocabulary
ordering. The rank is taken as the minimum across a band of source layers
(8--28 for Llama, 10--38 for gemma, 7--24 for Qwen; the band tracks
$d_{\mathrm{model}}$). Episodes are the arbitration case of the conflict
construction --- the model's closed-book answer is correct, the tool
return carries the typed distractor --- split by what the model actually
did, from the behavioural labels of the same runs. As a control we take,
for the same prompts, the elicited-answer rows of \emph{other} questions,
for which the model holds no such answer here. Table~\ref{tab:jlens} reports the ranks.

\begin{table}[t]
\centering
\setlength{\tabcolsep}{4pt}
{\small
\begin{tabularx}{\columnwidth}{Xccccc}
\toprule
 & \multicolumn{2}{c}{Followed the tool} & \multicolumn{2}{c}{Kept memory} & Control \\
\cmidrule(lr){2-3}\cmidrule(lr){4-5}\cmidrule(lr){6-6}
Model & Median & Top-20 & Median & Top-20 & Median \\
\midrule
Llama-3.1-8B-Instruct, seed 1 & 1 & 38/41 & 1 & 53/55 & 421 \\
Llama-3.1-8B-Instruct, seed 2 & 1 & 33/37 & 1 & 59/63 & 421 \\
gemma-2-9b-it                 & 2 & 32/36 & 1 & 36/39 & 538 \\
Qwen2.5-7B-Instruct           & 26{,}145 & 0/25 & 34{,}800 & 0/66 & 36{,}598 \\
\bottomrule
\end{tabularx}}
\caption{Rank of the model's own elicited closed-book answer in the
Jacobian-lens reading at the final input position, on the arbitration case, by
what the model then did. Lower is closer to the top of the distribution. The
control column applies the same measure to answers the model does not hold for
the question at hand, and is a per-model quantity, so the two Llama seeds share
it.}
\label{tab:jlens}
\end{table}

\textbf{Reading.} On both Llama seeds the model's own correct answer sits
at rank 1 whether or not the model goes on to produce it: adopting the
wrong tool value does not push the parametric answer down the
distribution, against a control median of 421. On gemma the pattern holds
with one asymmetry --- rank 1 when it keeps its answer, rank 2 when it
follows the tool. On Qwen there is nothing to see: the median rank is
around $2.6 \times 10^{4}$, no episode places the answer in the top 20,
and the control sits at the same order of magnitude. This is not an
artifact of how the answer is tokenized: scoring the best rank over all
subtokens of the answer span rather than its first row moves the Qwen
medians only to 25{,}548 and 27{,}378, still with nothing in the top 20,
while leaving the Llama numbers unchanged.

Read narrowly, the probe says that on two of four families the failure in
the arbitration case is not a failure of \emph{availability}. The answer
is present and near the top of the read-out at the moment the model writes
the tool's value instead. That is consistent with the behavioural picture
in \S\kcbsec{sec:tuning-results}{5.2} --- interventions can move the choice without teaching the model
anything new --- but it is a suggestion from a probe, not evidence for it.

\textbf{Four caveats, all load-bearing.} First, the rank is that of the
\emph{memory} answer --- the string the model produced under forced
closed-book elicitation --- and not of the gold answer as such; on this
cell the two coincide by construction, but the measure follows the
model's answer, and the distinction matters wherever they could differ.
Second, in a majority of the kept-memory episodes the answer string also
occurs somewhere in the prompt, where a top rank is much less surprising;
restricted to episodes where it does not, the counts are smaller but the
picture is unchanged (Llama seed 1: 25/26 in the top 20 after following
the tool, 11/11 after keeping memory; gemma 21/24 and 8/8; Qwen 0/7 and
0/7). Third, the effect is absent on Qwen and only partial on gemma, so it
is not a property of transformers here but of some of them. Fourth, a
lens read at one position is a measurement of the representation under a
particular projection, and we make no claim that it identifies a
mechanism.

\fi%

\end{document}